\documentclass[10pt,twocolumn,letterpaper]{article}

\usepackage[accsupp]{axessibility}  

\usepackage[pagenumbers]{cvpr} 
\usepackage{multirow}
\usepackage{booktabs}
\usepackage{array}
\usepackage{caption}
\usepackage{tcolorbox}
\usepackage{listings}
\usepackage[table]{xcolor}
\definecolor{mygray}{gray}{.6}
\definecolor{myblue}{RGB}{89,158,254}
\definecolor{mygreen1}{RGB}{81,150,111}
\definecolor{mygreen2}{RGB}{93,174,86}
\definecolor{myred}{RGB}{160,0,0}
\definecolor{myyellow}{RGB}{227,207,87}
\definecolor{cvprblue}{rgb}{0.21,0.49,0.74}
\usepackage[pagebackref,breaklinks,colorlinks,allcolors=cvprblue]{hyperref}

\def\name{Yo'City}

\def\eg{\emph{e.g.}}

\title{\name: Personalized and Boundless 3D Realistic City Scene Generation \\ via Self-Critic Expansion}

\author{
    Keyang Lu$^{1,2}$,
    Sifan Zhou$^{3}$,
    Hongbin Xu$^{4}$,
    Gang Xu$^{5}$,
    Zhifei Yang$^{1}$ \\
    Yikai Wang$^{6}$,
    Zhen Xiao$^{1*}$,
    Jieyi Long$^{7}$,
    Ming Li$^{5*}$ \\
    $^1$School of Computer Science, Peking University \quad
    $^2$Beihang University \quad
    $^3$Southeast University \\
    $^4$ByteDance Seed \quad
    $^5$Guangdong Laboratory of Artificial Intelligence and Digital Economy (SZ) \\
    $^6$Beijing Normal University \quad
    $^7$Theta Labs \\
}

\begin{document}
\twocolumn[{%
    \maketitle
    \centering
    \vspace{-6mm}
    \includegraphics[width=\textwidth]{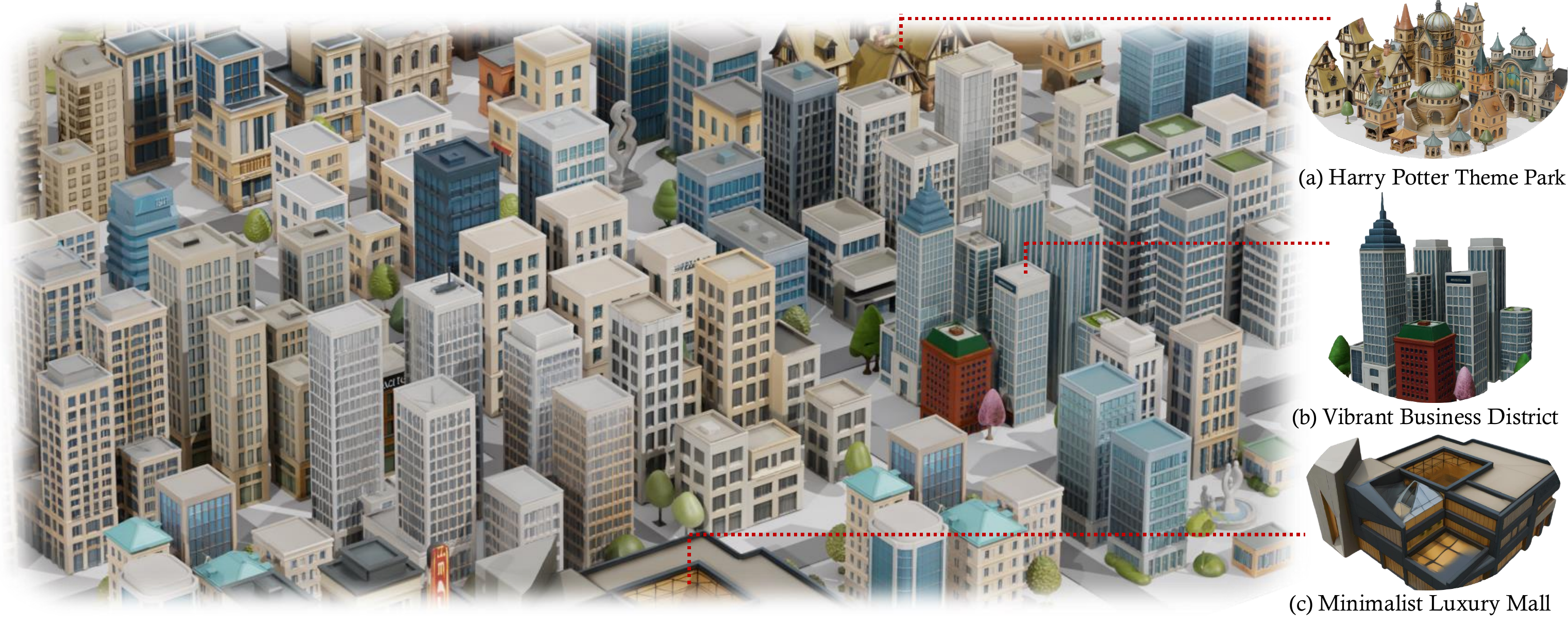}
    \captionof{figure}{
        \textbf{A vast city generated by Yo'City}. It incorporates key elements of a modern metropolis while also featuring more personalized designs, such as a Harry Potter–themed park and a minimalist shopping mall. The zoomed-in views of them are provided on the right.
    }
    \label{fig:teaser}
    \vspace{0.8em}
}]
\def\thefootnote{}\footnotetext{$\star$ Corresponding authors.} 
\begin{abstract}
Realistic 3D city generation is fundamental to a wide range of applications, including virtual reality and digital twins. However, most existing methods rely on training a single diffusion model, which limits their ability to generate personalized and boundless city-scale scenes.
In this paper, we present \textbf{\name{}}, a novel agentic framework that enables user-customized and infinitely expandable 3D city generation by leveraging the reasoning and compositional capabilities of off-the-shelf large models.
Specifically, \name{} first conceptualizes the city through a top-down planning strategy that defines a hierarchical “City–District–Grid” structure. The Global Planner determines the overall layout and potential functional districts, while the Local Designer further refines each district with detailed grid-level descriptions. 
Subsequently, the grid-level 3D generation is achieved through a “produce–refine–evaluate” isometric image synthesis loop, followed by image-to-3D generation.
To simulate continuous city evolution, \name{} further introduces a user-interactive, relationship-guided expansion mechanism, which performs scene graph–based distance- and semantics-aware layout optimization, ensuring spatially coherent city growth.
To comprehensively evaluate our method, we construct a diverse benchmark dataset and design six multi-dimensional metrics that assess generation quality from the perspectives of semantics, geometry, texture, and layout.
Extensive experiments demonstrate that \name{} consistently outperforms existing state-of-the-art methods across all evaluation aspects.


\end{abstract}    
\vspace{-5mm}
\section{Introduction}
\label{sec:intro}

3D city models play a crucial role in numerous applications, including virtual reality~\cite{ vincur2017vr}, gaming~\cite{tan2016evolution}, urban planning~\cite{shan2021research}, digital twins~\cite{schrotter2020digital}, and robotics~\cite{zhao2025resilient, ma2025ctsg}.
However, as urban environments consist of massive numbers of buildings with diverse heights, styles, and layouts, manually constructing such complex and large-scale scenes remains extremely challenging and labor-intensive.

Traditional approaches, such as procedural modeling~\cite{parish2001procedural, groenewegen2009procedural, talton2011metropolis} and image-based modeling~\cite{vezhnevets2007interactive, fan2014structure, aliaga2008interactive}, rely heavily on prior knowledge, \eg, handcrafted rules or street-view imagery, which limits their scalability and efficiency.
With the advent of generative models, recent works~\cite{shen2022sgam, lin2023infinicity, xie2024citydreamer, Deng_2025_CVPR, deng2024citycraft} have explored 3D city generation using GANs or diffusion models, where semantic layouts and height fields are generated and subsequently reconstructed into urban scenes.
Another line of research~\cite{wu2024blockfusion, lee2025nuiscene, hua2025sat2city} focuses on volumetric latent representations, aiming to synthesize urban environments in a compact and geometrically consistent manner.
However, these methods usually require maps or satellite datasets for training and struggle to handle flexible, user-friendly text inputs.

With the rapid advancement of large language models (LLMs) and vision-language models (VLMs), agentic frameworks have been widely adopted across diverse domains like scientific research~\cite{seo2025paper2code, yue2025foam, koh2025c2} and multimodal reasoning~\cite{wu2025vtool,su2025openthinkimg, huang2025visualtoolagent, 10.1145/3581783.3611756,fu2025mano}.
Benefiting from their rich world knowledge and powerful perception, reasoning, and planning capabilities, these agents can execute multi-step solutions that involve external tools and complex decision-making—tasks that are typically infeasible for a conventional single model. Similarly, several studies have explored 3D indoor scene synthesis guided by LLMs or VLMs~\cite{feng2023layoutgpt, ccelen2024design, sun2025layoutvlm, yang2024holodeck}. 
However, agentic 3D city generation remains largely unexplored. Unlike closed indoor environments, cities are open, large-scale, and highly structured spaces containing a far greater diversity of objects and much denser spatial organization, thereby posing significant challenges for realistic and scalable 3D scene generation.

A recent work, SynCity~\cite{Engstler_2025_ICCV}, explores training-free 3D scene generation through an autoregressive tile-by-tile pipeline, combining a 2D image generator and a 3D modeler via prompt engineering. Each tile is generated sequentially and fused with previously synthesized tiles to form a complete scene, yet the framework lacks an explicit planning mechanism to reason about urban structure and spatial hierarchy. However, this flat generation paradigm does not align with the intrinsic organization of real-world cities, which typically exhibit \textit{a distinct hierarchical structure}—each district can be subdivided into functional blocks that maintain internal coherence while remaining spatially connected to others. As a result, SynCity performs well on small or locally coherent scenes, but struggles to maintain global consistency when scaled to large, realistic city environments.
Moreover, the absence of hierarchical reasoning and refinement mechanisms leads to simplified geometry, cartoonish appearance, and blurry textures, ultimately resulting in low realism for city-scale synthesis.



In this paper, we propose \name{}, a multi-agent framework for boundless and realistic 3D city generation driven by user-customized text inputs. Inspired by the hierarchical logic ``City–District–Grid'' of real-world cities, we design a coarse-to-fine thinking strategy that enables both global structural planning at the city level and fine-grained architectural design at the grid level. Concretely, the Global Planner acts as a high-level controller that interprets user intent, analyzes urban functions, and allocates districts on a grid map with estimated sizes and adjacency.
At the district level, the Local Designer refines these blueprints into grid-level descriptions, defining architectural styles, building densities, landmarks, and surrounding context.
This thinking strategy allows \name{} to reason globally while designing locally with rich geometric and semantic details.


Based on the grid descriptions, we design a produce–refine–evaluate isometric image synthesis loop that preserves spatial consistency while enhancing architectural diversity. The generated isometric images are then converted into 3D assets via an image-to-3D generator, followed by post-processing to assemble them into a coherent urban scene. To enable continuous city evolution, \name{} further introduces a scene-graph–based self-critic expansion module. Given user preferences, the module automatically infers the structure of the new grid and builds a scene graph encoding their relationships with existing districts.
A distance- and semantics-aware optimization is then applied to determine the most plausible placement. Through this mechanism, \name{} achieves unbounded and spatially coherent city generation. We establish a multi-dimensional evaluation benchmark to comprehensively assess our framework. \name{} consistently outperforms prior approaches in VQAScore~\cite{lin2024evaluating}, geometric fidelity, layout coherence, texture clarity, scene coverage, and overall realism.

Our main contributions are summarized as follows:
\begin{itemize}
\item We propose \name{}, a novel multi-agent framework for boundless and realistic 3D city generation guided by user-customized textual instructions.
\item We model the city with a grid-based hierarchical structure and design a top-down planning strategy to generate spatially coherent urban layouts. To enable plausible and automated city expansion, we further introduce a scene graph–based mechanism that performs distance- and semantics-aware location optimization.
\item We construct a multi-dimensional evaluation benchmark that assesses semantic consistency and visual quality in five aspects—geometric fidelity, texture clarity, layout coherence, scene coverage, and overall realism.
\end{itemize}

\section{Related Work}
\subsection{3D City Generation}
3D scene generation aims to create high-quality 3D environments based on various types of input, including both indoor~\cite{paschalidou2021atiss, tang2024diffuscene, yang2024holodeck, yang2025mmgdreamer,Yang_2024_CVPR} and outdoor scenes~\cite{yu2025wonderworld,zhou2025scenex,Chai_2023_CVPR,10269790}. Among them, 3D cities have become a key research focus due to their complex layouts and diverse architectural forms. Classic methods often rely on manually defined rules~\cite{  kelly2021cityengine} and image-based techniques~\cite{vezhnevets2007interactive, fan2014structure, aliaga2008interactive}, or perform procedural modeling~\cite{ghorbanian2019procedural, groenewegen2009procedural, talton2011metropolis, lipp2011interactive} through simulation engines, all of which can be inefficient and inflexible. Most current approaches~\cite{lin2023infinicity, xie2024citydreamer, Deng_2025_CVPR, deng2024citycraft, niu2025controllable} first obtain a 2D semantic map of the city using generative models, and then generate individual buildings based on this map through retrieval or generative methods, forming a city scene. With the development of 3D representation, some studies employ diffusion models to directly generate large-scale scenes in 3D space, such as~\cite{wu2024blockfusion, lee2025nuiscene, hua2025sat2city, lee2025skyfall}. However, the aforementioned methods often require extensive real-world data (like maps or  satellite images) and lack intuitive user controls, making them unsuitable for personalized requirements and difficult to expand to large-scale scenes through interactions.

\subsection{Agentic Systems}
The rapid advancement of large language models~\cite{floridi2020gpt,touvron2023llama,anil2023palm} and vision-language models~\cite{bai2025qwen3,achiam2023gpt, liu2023visual} has significantly enhanced the capabilities of agents. By leveraging the abundant knowledge of these models and their powerful understanding abilities, agents empower various fields including software engineering~\cite{qian2023chatdev, yang2024swe, zhang2024codeagent}, visual understanding~\cite{li2023llava, kim2024image, Fang_2023_ICCV, yang2025vidtext} and spatial perception~\cite{pan2025metaspatial, yin2025spatial, pan2025metafind, 11208990}, etc. Many works have also applied agents to 3D scene generation, especially for indoor scene generation~\cite{feng2023layoutgpt, ccelen2024design, sun2025layoutvlm, yang2024holodeck, gu2025artiscene}. Some studies~\cite{yao2025cast, ling2025scenethesis, bian2025holodeck} also introduce training-free and user-friendly approaches for outdoor scene synthesis, but they typically rely on a single reference image, posing challenges for the generation of vast urban environment. Recently, SynCity~\cite{Engstler_2025_ICCV} improves it by employing a tile-by-tile pipeline. Nevertheless, it shows unsatisfactory performance in generating large-scale city scenes with dense structures, and the results are lack of realism and fidelity. We solve these drawbacks based on our proposed agentic framework.
\vspace{-4.9mm}
\section{Yo'City}
\begin{figure*}
    \centering
    \includegraphics[width=\linewidth]{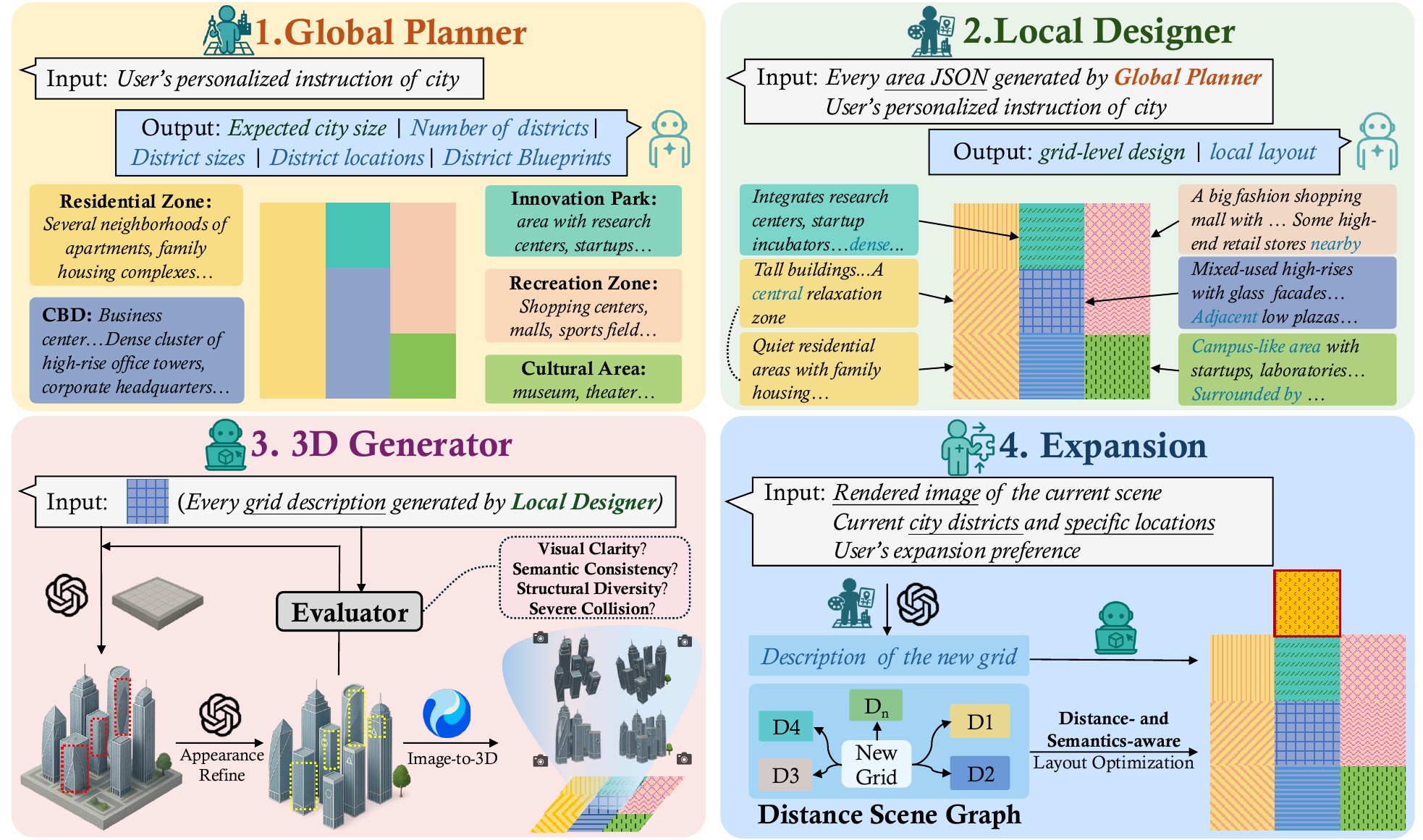}
    \caption{\textbf{Overview of Yo'City}. Global Planner: Converts the user prompt into a coarse city layout. Local Designer: Refines the layout into detailed, per-grid textual descriptions. 3D Generator: Synthesizes 3D assets for each grid by lifting isometric images. Expansion Module: Determines the content and optimal placement for new grids to evolve the city. Finally, all generated 3D assets are assembled into the complete city scene.
    } 
    \label{fig:pipeline}
    \vspace{-5mm}
\end{figure*}
\subsection{Problem Formulation}
We define personalized and boundless 3D city generation as a “planning-generation-expansion” task. Given an arbitrary textual prompt $p_0$ describing the user's preferences for a city, our goal is to generate a well-planned and realistic 3D city model $\mathcal{G}$ that can subsequently evolve. We formulate this by spatially partitioning the world into a $H\times W$ grid of tiles, denoted by $\mathcal{T} = \{0, \ldots, H-1\} \times \{0, \ldots, W-1\}$, where each tile $(x,y) \in \mathcal{T}$ is a 3D scene patch (e.g., a residential community) and the underlying ground surface. Crucially, unlike auto-regressive methods~\cite{Engstler_2025_ICCV} (tile-by-tile) synthesis, our \name{} generates all tiles in $\mathcal{T}$ in parallel. We eliminate the strict causal dependency where the generation of tile $(x,y)$ must be conditioned on the set of previously generated tiles $\mathcal{T}(x,y)$. This parallel formulation breaks the causal dependency on prior tiles $\mathcal{T}(x,y)$, which not only significantly accelerates the generation but also avoids the potential error accumulation inherent in sequential processes. Our goal is thus able to generate the properties of all tiles in $\mathcal{T}$ simultaneously, based on the global prompt $p_0$. 




\subsection{Framework Overview}
As illustrated in Fig.~\ref{fig:pipeline}, \name{} framework systematically generates and expands a 3D city with four key modules: Global Planner, Local Designer, 3D Generator and Expansion Module. First, the \textit{Global Planner} (Sec.~\ref{sec:global_planner}) translates high-level user prompt $p_0$ into an overall urban layout, defining the number, size, functions and locations of districts. Second, the \textit{Local Designer} (Sec.~\ref{sec:local_designer}) refines the district design into detailed, grid-level textual descriptions, culminating in a comprehensive city blueprint through hierarchical coarse-to-fine planning. Third, based on the per-grid descriptions, the \textit{3D Generator} (Sec.~\ref{sec:3d_generator}) synthesizes each grid by first generating a 2D isometric image via a produce-refine-evaluate loop, and subsequently lifting it to a 3D asset using a pretrained model~\cite{lai2025hunyuan3d}. By arranging all grids according to the previously generated layout, \name{} achieves a well-structured and realistic 3D city scene. Finally, the \textit{Relationship-guided Expansion} module (Sec.~\ref{sec:expansion}) enables city evolution. It employs a VLM and a graph-based optimization to adaptively determine the optimal placement for the new tile, which is then synthesized and seamlessly integrated.


\subsection{Global Planner}
\label{sec:global_planner}
The Global Planner, as illustrated in Fig.~\ref{fig:pipeline}.1, translates the abstract and personalized user prompt $p_0$ into a high-level city layout. We first model the city using a hierarchical “City–District–Grid” structure. The planning process is three-fold: \textbf{(i) Size Estimation}: an LLM first estimates the city size, represented as a rectangular grid of $H \times W$, where $H$ and $W$ correspond to the total numbers of rows and columns, respectively. Each grid cell can be regarded as a block and serves as the fundamental spatial unit within the urban layout. \textbf{(ii) District Planning}: The planner then identifies $N$ distinct functional districts and generates a set of concise blueprints $\{B_i \mid i = 1, 2, \dots, N\}$, where each $B_i$ represents the conceptual plan of the $i$-th district and $N$ denotes the number of districts. \textbf{(iii) Layout Allocation}: Taking potential spatial relationships and proximity constraints among different regions into account, these districts are coherently allocated onto the predefined $H \times W$ grid map, which may span multiple grid cells to cover various areas. Each $B_i$ specifies the district’s function (e.g., “business center”) and its constituent building types (e.g., “high-rise office towers”), providing the global structural priors for the subsequent Local Designer (Sec.~\ref{sec:local_designer}).


While methods like SynCity~\cite{Engstler_2025_ICCV} employ manually-constructed instructions or rely on LLM's intrinsic knowledge to interpret high-level prompts,
they often lack factual grounding for specific real-world references (e.g., “New York-like”). To address this limitation, we further introduce a Retrieval-Augmented Generation (RAG) module. This module retrieves relevant information about the reference city from a curated Wikipedia corpus, focusing on its urban structure, zoning characteristics, and spatial organization patterns. The retrieved content is then distilled by the GPT-4o-mini model~\cite{achiam2023gpt}, extracting representative structural and functional traits, which are then integrated as prior knowledge into the global planning process. This allows the generated city to better align with the spatial logic and aesthetic characteristics of the referenced real-world city, while maintaining flexibility for personalization and creativity.


\subsection{Local Designer}
\label{sec:local_designer}
Building on the blueprints $\{B_i\}$ generated by the Global Planner (Sec.~\ref{sec:global_planner}), we develop a Local Designer to refine these coarse plans into fine-grained, grid-level descriptions. Specifically, for each district, the LLM is conditioned on both its blueprint $B_i$ and the global user prompt $p_0$. It then generates detailed designs for all grids within the city, denoted as$\{ d_i \mid i = 1, 2, \ldots, H \times W \}$, where each $d_i$ is a textual representation capturing the target grid's spatial organization and visual characteristics. Crucially, to ensure continuity, the Local Designer jointly plans for all grids within a multi-grid district, enforcing spatial and stylistic coherence across them. Compared to generating the entire city layout in a single step, our coarse-to-fine strategy provides the LLM with an implicit reasoning process. By decomposing the task into global and local stages, the model reasons progressively—from high-level organization to fine-grained details, enabling more structured planning and getting layouts with greater realism, consistency, and plausibility.



\subsection{3D Generator}
\label{sec:3d_generator}
The 3D Generator lifts the grid-level descriptions $\{d_i\}$ into 3D assets. This process involves two stages: (1) generating a high-quality 2D isometric image for each grid as an intermediate representation, and (2) converting these images into 3D models.

\noindent\textbf{2D Isometric Image Generation.} A naive text-to-image approach for stage (1) is insufficient, as it often produces misaligned objects or partial views with incomplete buildings, failing to maintain inter-grid spatial consistency. 
While static constraints like a fixed base can enforce an isometric perspective, they lack fine-grained quality control. 
For example, some grids may contain an excessive number of buildings, while others appear overly sparse. We therefore introduce an iterative produce–refine–evaluate loop to ensure both structural coherence and high fidelity: \textbf{(i) Produce}: We first generate an initial isometric image for the grid $d_i$ on a pre-defined ground platform. This platform acts as a common anchor, ensuring all generated assets share a consistent scale and spatial alignment. \textbf{(ii) Refine}: An image editing model then removes this platform and refines the asset's surfaces, correcting potential geometric artifacts and enhancing visual diversity. \textbf{(iii) Evaluate}: A specialized evaluator assesses the refined image for text-image alignment, realism, and layout rationality. The feedback is passed back for generation until all quality criteria are met.

\noindent\textbf{3D Model Conversion.} The resulting high-quality and coherent isometric images are then converted into 3D models using a pretrained image-to-3D model~\cite{lai2025hunyuan3d}.

\noindent\textbf{Scene Assembly.} After generating 3D models for all grids, we assemble the final city. Leveraging our parallel, grid-aligned generation pipeline, the 3D models can be directly arranged according to the predefined layout from the Global Planner (Sec.~\ref{sec:global_planner}) without requiring complex 3D blending to resolve boundary inconsistencies. We then add essential elements, such as roads and ground surfaces, to connect the grids. This assembly stage is style-aware and ground materials and other attributes can also be customized by users to match the city's theme (e.g., ancient or modern).

\subsection{Relationship-guided Expansion}
\label{sec:expansion}
In urban systems, the spatial proximity principle dictates how functional regions are organized and interact with one another. Certain areas need to be spatially coordinated to achieve both accessibility and harmony within the city. For example, residential zones are generally positioned near schools and business zones to support everyday convenience and commute, while industrial districts are deliberately planned at a greater distance to avoid conflicts caused by noise or pollution. To incorporate such distance-based spatial constraints, we introduce a relationship-guided expansion mechanism (Fig.~\ref{fig:pipeline}.4). 

This process begins when a user provides an expansion demand. Given the rendered city and a regions overview, a VLM performs two key tasks: (i) it reasons over the existing scene to generate a textual description for the target expansion grid, $d_\text{new}$, and (ii) it constructs a scene graph capturing the potential relationships between the new grid and the existing districts. In the scene graph, the new grid acts as the central node, with edges to existing districts encoding qualitative distance relationships (e.g., "near", "relatively near"). Based on it, we design a distance- and semantics-aware optimization function that integrates both spatial relationship reasoning and semantic coherence, which is applied to determine the most suitable position for the new expansion grid. The goal is to select a feasible grid location that best satisfies the distance relationships inferred by the VLM while maintaining contextual harmony with the adjacent grids.



Formally, we first employ breadth-first search over the city layout to identify a set of feasible candidate locations $\mathcal{X}$. Let $\mathcal{G} = \{ g_1, g_2, \dots, g_{H\times W} \}$ represent all grids in the existing city. 
Each grid $g_i$ within a district is associated with a qualitative spatial relationship $r(g_i) \in \{\text{near}, \text{relatively near}, \text{slightly near}, \text{no special constraint}, \\ \text{far}\}$ derived from the scene graph, and a corresponding weight $\gamma_{r(g_i)}$ quantifying the relative importance of the relationship type.

\vspace{0.5em}
\noindent\textbf{Distance-driven Spatial Objective.}  
For each candidate location $x \in \mathcal{X}$, we compute its Euclidean distance to other grids $g_i$. 
The spatial objective aggregates these distances, weighted by the qualitative relationship weight:
\begin{align}
\label{eq:distance_term}
L_{\text{dist}}(x) = 
\sum_{g \in \mathcal{G}} 
\gamma_{r(g)} \,  \|x - g\|_2 ,
\end{align}
where $\|x - g\|_2$ denotes the Euclidean distance between the candidate grid $x$ and the existing grid $g$, and $\gamma_{r(g)}$ are signed: positive for proximity (pulling $x$ closer) and negative for separation (pushing $x$ away).
This term enforces the spatial coordination, guiding the expansion grid to be closer to regions that should maintain strong spatial relations and farther from those that should remain separated.

\vspace{0.5em}
\begin{table*}[ht]
\centering
\small
\caption{\textbf{Quantitative comparison of different methods across six evaluation dimensions.} We use the VQAScore to evaluate semantic consistency. For the five aspects of visual quality, we conduct pairwise comparisons evaluated by both GPT-5 and human judges, and reported the win rate for each method. To reduce randomness, each comparison is performed twice.}
\label{tab:quantitative_comparison}
\setlength{\tabcolsep}{2.8pt}
\renewcommand{\arraystretch}{1.15}
\resizebox{\textwidth}{!}{{
\begin{tabular}{>{\centering\arraybackslash}m{3cm} >{\centering\arraybackslash}p{1.4cm} cc cc cc cc cc}
\toprule
\multirow{2}{*}{\textbf{Method}} & \multirow{2}{*}{\textbf{VQAScore}}
& \multicolumn{2}{c}{\textbf{Geometric Fidelity}} 
& \multicolumn{2}{c}{\textbf{Texture Clarity}} 
& \multicolumn{2}{c}{\textbf{Layout Coherence}} 
& \multicolumn{2}{c}{\textbf{Scene Coverage}} 
& \multicolumn{2}{c}{\textbf{Overall Realism}} \\
\cmidrule(lr){3-4} \cmidrule(lr){5-6} \cmidrule(lr){7-8} \cmidrule(lr){9-10} \cmidrule(lr){11-12}
& & GPT-5 & Human & GPT-5 & Human & GPT-5 & Human & GPT-5 & Human & GPT-5 & Human \\
\midrule
Trellis~\cite{xiang2025structured} & 0.6189 & 6.50\% & 7.00\% & 4.50\% & 6.00\% & 6.50\% & 3.50\% & 6.00\% & 3.50\% & 9.00\% & 5.00\% \\
\rowcolor{myblue!18} \textbf{Yo'City (Ours)} & 0.7151 & 93.50\% & 93.00\% & 95.50\% & 94.00\% & 93.50\% & 96.50\% & 94.00\% & 96.50\% & 91.00\% & 95.00\% \\
\midrule
Hunyuan3D (API)~\cite{lai2025hunyuan3d} & 0.6198 & 12.00\% & 7.00\% & 12.50\% & 9.50\% & 7.00\% & 5.50\% & 3.50\% & 4.00\% & 12.00\% & 6.50\% \\
\rowcolor{myblue!18} \textbf{Yo'City (Ours)} & 0.7151 & 88.00\% & 93.00\% & 87.50\% & 90.50\% & 93.00\% & 94.50\% & 96.50\% & 96.00\% & 88.00\% & 93.50\% \\
\midrule
CityCraft~\cite{deng2024citycraft} & 0.5639 & 9.50\% & 8.00\% & 6.00\% & 6.00\% & 15.00\% & 16.50\% & 23.50\% & 25.00\% & 12.00\% & 13.50\% \\
\rowcolor{myblue!18} \textbf{Yo'City (Ours)} & 0.7151 & 90.50\% & 92.00\% & 94.00\% & 94.00\% & 85.00\% & 83.50\% & 76.50\% & 75.00\% & 88.00\% & 86.50\% \\
\midrule
SynCity~\cite{Engstler_2025_ICCV} & 0.6975 & 15.00\% & 12.00\% & 21.50\% & 18.50\% & 14.00\% & 10.50\% & 18.00\% & 15.50\% & 15.50\% & 12.00\% \\
\rowcolor{myblue!18} \textbf{Yo'City (Ours)} & 0.7151 & 85.00\% & 88.00\% & 78.50\% & 81.50\% & 86.00\% & 89.50\% & 82.00\% & 84.50\% & 84.50\% & 88.00\% \\
\bottomrule
\end{tabular}
}}
\vspace{-2mm}
\end{table*}

\noindent\textbf{Semantic Regularization.}  
While the distance term captures layout relationships, it does not guarantee that the new grid is compatible with its surrounding context. 
To ensure semantic coherence, we introduce a semantic regularization term based on the Sentence-Bert~\cite{reimers2019sentence} embedding similarity between $d_\text{new}$ and its neighboring grids $\mathcal{N}(x)$.
\begin{equation}
\label{eq:semantic_term}
L_{\text{sem}}(x) = 
-\!\!\!\sum_{y \in \mathcal{N}(x)}
\text{Embedding\_Sim}\!\left(d_\text{new}, d_y\right),
\end{equation}
A higher embedding similarity indicates better semantic compatibility; hence, this term encourages selecting a location where the new grid can blend naturally into the existing urban context.

\vspace{0.5em}
\noindent\textbf{Overall Objective.}  
Combining the spatial and semantic components, the final optimization objective is defined as:
\begin{equation}
\label{eq:total_objective}
L(x) = L_{\text{dist}}(x) + \lambda \, L_{\text{sem}}(x),
\end{equation}
where $\lambda$ balances the contribution of semantic regularization. 
The optimal expansion position is obtained by minimizing Eq.~(\ref{eq:total_objective}):
\vspace{-2mm}
\begin{align}
x^* = \arg\min_{x \in \mathcal{X}} L(x).
\label{eq:argmin}
\end{align}
After obtaining the optimal placement location $x^*$, we utilize the 3D generator (Sec.~\ref{sec:3d_generator}) to synthesize the corresponding 3D model of the new grid $ d_{\text{new}}$, completing the process. Empowered by this relationship-guided expansion, \name{} enables iterative expansion of the generated city through user interactions, supporting truly open-world and boundless generation.

\vspace{-1mm}
\section{Experiments}
\subsection{Settings}
\noindent\textbf{Dataset}. To evaluate our method, we construct a dataset of 100 textual descriptions of cities, of which 30\% are manually written and 70\% are generated by the GPT-4o model. The dataset contains various types of textual descriptions, which is elaborated in Appendix.~\ref{dataset_curation}.

\noindent\textbf{Baselines}. For comparison, we adopt Trellis~\cite{xiang2025structured}, Hunyuan3D (API)~\cite{lai2025hunyuan3d}, CityCraft~\cite{deng2024citycraft} and SynCity~\cite{Engstler_2025_ICCV} as baseline methods. 
Trellis and Hunyuan3D are widely used and representative text-to-3D generation models. CityCraft is a learning-based method that first generates a scene layout and then retrieves predefined assets to populate it, whereas SynCity is a recently proposed training-free autoregressive approach for large-scale world generation.

\noindent\textbf{Implementation details}. In our experiments, we adopt GPT-4o as the large language model, GPT-Image-1 for image editing, and Hunyuan3D (API) for image-to-3D generation. The specific experimental setup and hyperparameter configurations are provided in Appendix.~\ref{implementation}.
\begin{figure*}
    \centering
    \includegraphics[width=\linewidth]{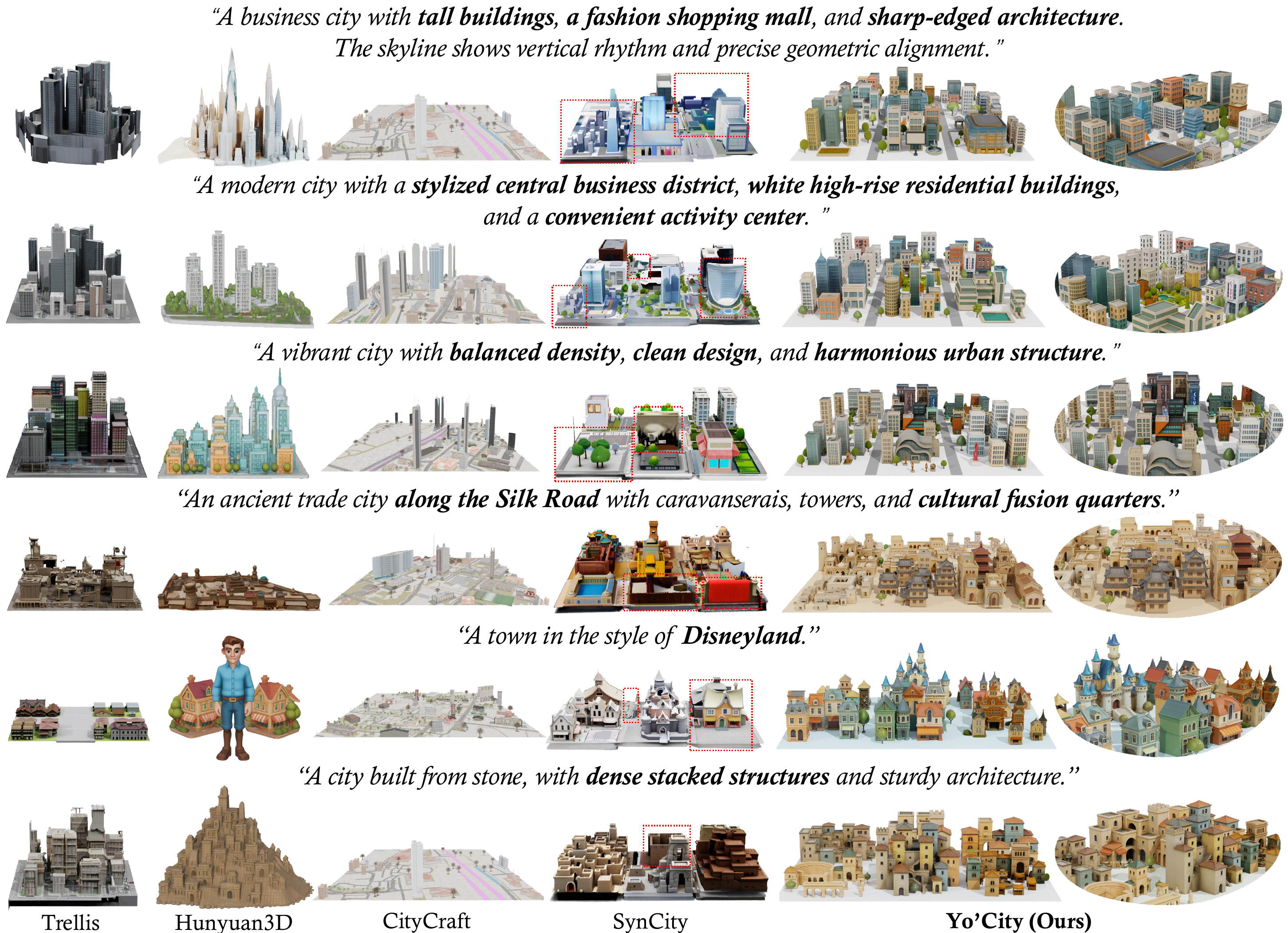}
    \caption{\textbf{Qualitative comparison between our method and the baselines given the same city instructions}. The red boxes highlight regions in SynCity that exhibit spatial inconsistency, lack of realism, and poor texture fidelity. We additionally provides zoom-in visualizations for Yo'City, demonstrating clearer structural coherence and finer visual details. More cases are shown in Appendix.~\ref{Supplementary_Qualitative_Comparison}.}
    \label{fig:case}
    \vspace{-4mm}
\end{figure*}
\subsection{Evaluation Metrics}
To provide a comprehensive assessment of diverse methods, we adopt multi-dimensional evaluation metrics as follows:

\noindent\textbf{Semantic Consistency}. We employ VQAScore~\cite{lin2024evaluating} to measure the semantic consistency between city instructions and generated scenes.

\noindent\textbf{Visual Quality.}  
To assess the visual performance of the generated scenes, we conduct a perceptual evaluation based on five aspects:
\textbf{\textit{Geometric Fidelity}}, \textbf{\textit{Texture Clarity}}, \textbf{\textit{Layout Coherence}}, \textbf{\textit{Scene Coverage}}, and \textbf{\textit{Overall Realism}}. 
Each aspect is evaluated through pairwise comparisons by GPT-5~\cite{singh2025openai} and ten human judges, with each comparison repeated twice to mitigate randomness. Detailed evaluation criteria are provided in Appendix.~\ref{evaluation}.

\subsection{Main Results}
\paragraph{Quantitative Comparison.}
Tab.~\ref{tab:quantitative_comparison} presents a quantitative comparison of different methods, reporting their VQAScores and win rates in pairwise visual quality evaluations conducted by GPT-5 and human judges. 
Given the same input, our model achieves the highest VQAScore, indicating its stronger ability to generate scenes that better align with users' personalized requirements. In addition to achieving superior text consistency, our coarse-to-fine planning and delicately-designed 3D generation strategy also lead to better visual quality, achieving win rates of at least 85.00\% in Geometric Fidelity, 86.00\% in Layout Coherence, 78.50\% in Texture Clarity, and 84.50\% in Overall Realism.
\vspace{-3mm}
\paragraph{Qualitative Comparison.}
As shown in Fig.~\ref{fig:case}, Yo'City consistently outperforms the baselines, producing well-proportioned buildings with clear textures and high-fidelity details (e.g., windows), while maintaining a coherent layout with consistent scales and appropriate spacing.
In contrast, both SynCity and CityCraft exhibit clear limitations. CityCraft generalizes poorly beyond modern-city prompts and often fails to follow non-modern or stylized descriptions. SynCity shows clear spatial inconsistency: as illustrated in the first and second rows, it generates a dense cluster of buildings in the lower-left tile while other tiles remain relatively sparse, resulting in an imbalanced spatial distribution. Its results also suffer from coarse textures.
Furthermore, Yo'City demonstrates strong capabilities in personalized generation, effectively modeling fine-grained cues such as “sharp-edged”, “Silk Road” and “stacked structures”. 

\begin{table}[t]
\centering
\renewcommand{\arraystretch}{0.95}
\caption{\textbf{Grid-level experimental comparison between SynCity and Yo’City}. We report the Alignment Score and Aesthetic Score for both methods for comprehensive assessment.}
\resizebox{0.9\linewidth}{!}{
\begin{tabular}{lcc}
\toprule
\textbf{Method} & \textbf{Alignment Score} & \textbf{Aesthetic Score} \\
\midrule
SynCity~\cite{Engstler_2025_ICCV} & 0.6572 & 4.95 \\
\rowcolor{myblue!18} \textbf{Yo'City (Ours)} & 0.6927 & 5.52 \\
\bottomrule
\end{tabular}
}
\label{tab:grid_scores}
\vspace{-5mm}
\end{table}

\begin{figure*}
    \centering
    \includegraphics[width=\linewidth]{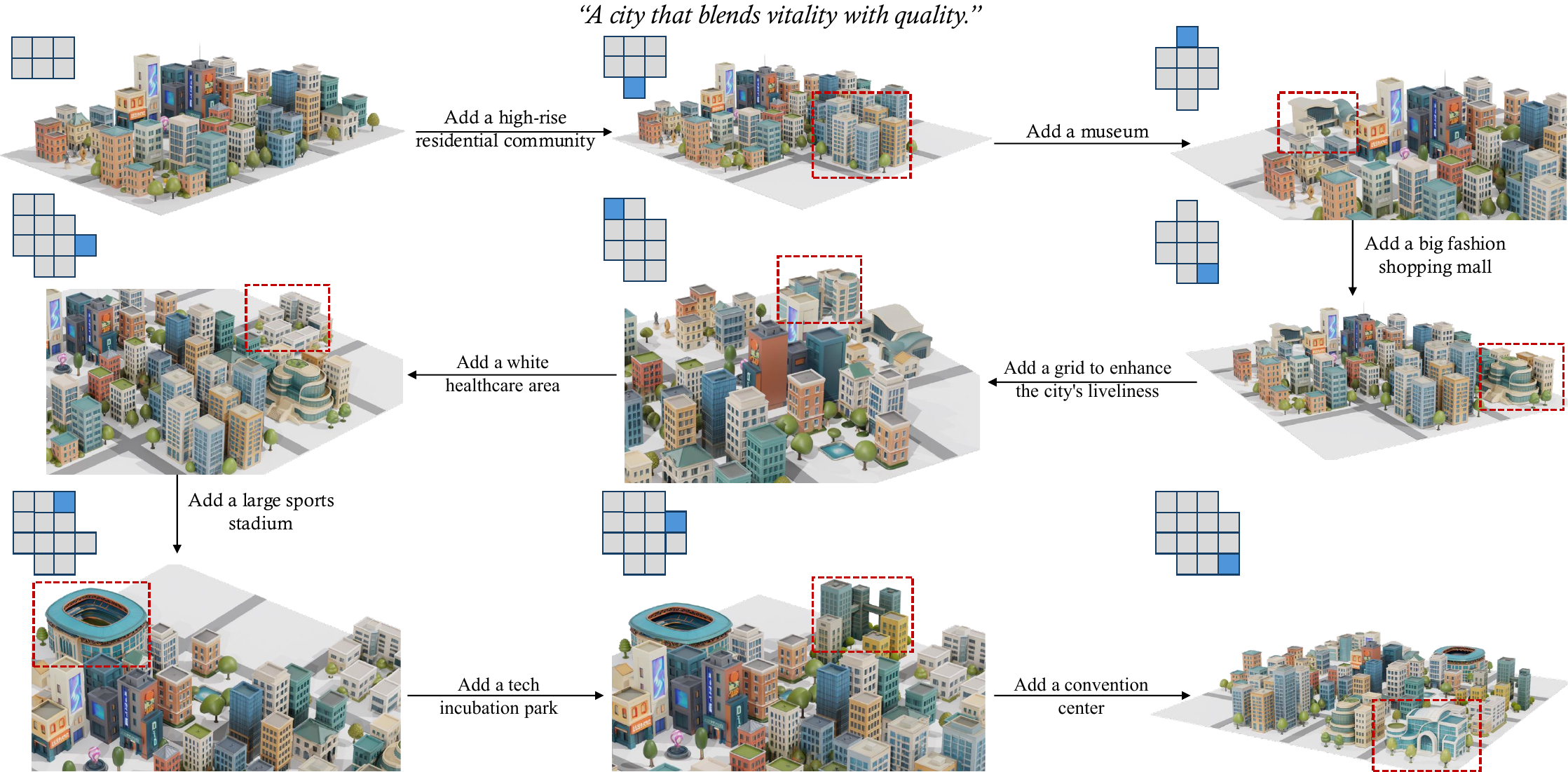}
    \vspace{-5mm}
    \caption{\textbf{Visualization of expansion}. The first row presents the city’s global instruction. The leftmost city shows the initial generation result, followed by five successive expansion iterations. In the top-left corner, a BEV thumbnail depicts the city layout, with blue regions indicating newly expanded grids, while red boxes in the rendered images highlight their appearances.
    } 
    \label{fig:ablation_expansion}
    \vspace{-4mm}
\end{figure*}
\vspace{-3mm}
\paragraph{Grid-level Experiment.}
\begingroup
\renewcommand{\thefootnote}{\fnsymbol{footnote}}

To further evaluate the structural and perceptual quality of generated city scenes, we conduct a grid-level experiment between SynCity and Yo'City. 
First, we measure whether each grid region semantically aligns with the target prompt by calculating the Alignment Score, defined as the VQAScore between the grid image and the query ``Does this figure show a reasonable grid of \{city prompt\}?''. 
This metric reflects the model’s ability to maintain consistent city-related semantics across different regions. 
Second, we assess the Aesthetic Score (with a full mark of 10) of each grid using an aesthetic predictor\hyperlink{fn:aesthetic_predictor}{\textsuperscript{\dag}}, which captures the visual appeal and scene fidelity of each grid.
As summarized in Tab.~\ref{tab:grid_scores}, Yo'City shows a clear advantage at the grid level (Alignment Score +0.0355, Aesthetic Score +0.57), being not only more consistent with the global instruction but also superior in overall aesthetics.
\footnotetext[2]{\hypertarget{fn:aesthetic_predictor} The aesthetic predictor is available at: \url{https://github.com/discus0434/aesthetic-predictor-v2-5}}

\endgroup
\subsection{Ablation Studies}
\begin{table}[t]
\centering
\small
\setlength{\tabcolsep}{3pt} 
\renewcommand{\arraystretch}{0.9}
\caption{\textbf{Ablation study on the planning strategy}. Yo’City (w/o reason) denotes the model without this strategy, while Yo’City (w reason) includes it. We evaluate both variants using VQAScore and GPT-5 win rates for Layout Coherence and Overall Realism.}
\label{tab:ablation_plan}
\begin{tabular}{lcc}
\toprule
\textbf{Metric} & \textbf{Yo'City (w/o reason)} & \textbf{Yo'City (w reason)} \\
\midrule
 VQAScore         &   0.7034     & \cellcolor{myblue!18} 0.7151       \\
Layout Coherence &   27.00\%    &  \cellcolor{myblue!18} 73.00\%      \\
Overall Realism  &   24.50\%     & \cellcolor{myblue!18} 75.50\%       \\
\bottomrule
\end{tabular}
\vspace{-3mm} 
\end{table}

\paragraph{Coarse-to-fine Planning.}
In Yo'City, we adopt a coarse-to-fine planning framework that enables coherent reasoning and decision-making during city generation. To evaluate its effectiveness, we compare it to a single-stage city planner (Yo'City w/o reason) that generates urban layouts directly from inputs. Both models are tested with the same one-shot example for fairness. As shown in Tab.~\ref{tab:ablation_plan}, results reveal that Yo'City (w. reason) outperforms Yo'City (w/o reason) in VQAScore, Layout Coherence, and Overall Realism, as evaluated by GPT-5. This improvement is attributed to the Global Planner and Local Designer, which better capture user preferences and produce a more organized city layout.

\vspace{-3mm}
\paragraph{Expansion Mechanism.}
\begin{figure}
    \centering
    \includegraphics[width=\linewidth]{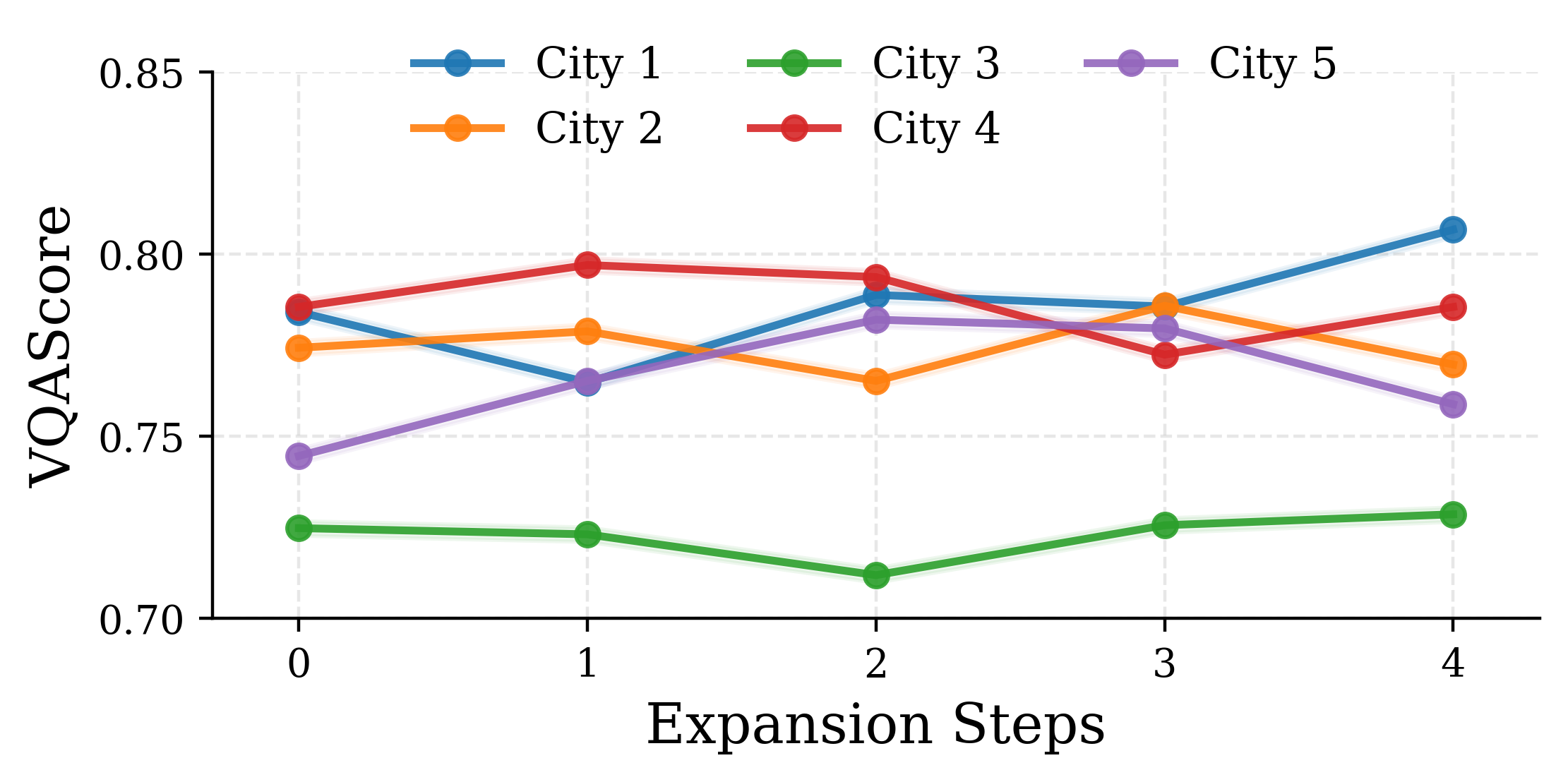}
    \vspace{-6mm}
    \caption{\textbf{VQAScore across four expansion steps} for five cities.}
    \label{fig:vqa_stability}
    \vspace{-6mm}
\end{figure}
To verify the effectiveness and robustness of our expansion mechanism, we select several city-level instructions and design four expansion tasks for each. After each expansion, we compute its VQAScore with respect to the corresponding global city instruction. During successive expansions, the generated city’s VQAScores stay stable, showing a Coefficient of Variation of 3.34\%. Fig.~\ref{fig:vqa_stability} shows the line chart of five experimental results and Fig.~\ref{fig:ablation_expansion} presents a larger-scale example with eight expansion steps. Given user preferences, Yo’City critiques the existing city to generate grid descriptions aligned with the global style and optimize the placement of them. For instance, shopping malls and healthcare areas are placed near residential neighborhoods for convenient access.
\vspace{-1mm}
\section{Conclusion}
In this paper, we propose Yo’City, a text-driven agentic framework for personalized and realistic 3D city generation without relying on map data. An LLM-based hierarchical planner designs city layouts, while a specialized 3D generator produces scale-consistent, detail-rich isometric images via a produce–refine–evaluate loop and converts them into 3D models. We further introduce a relationship-guided expansion mechanism that enables iterative city growth through textual instructions. Extensive experiments show that Yo’City outperforms existing methods across all dimensions, highlighting its potential for applications such as virtual reality and simulation games.


{
    \small
    \bibliographystyle{ieeenat_fullname}
    \bibliography{main}

@String(CVPR= {IEEE Conf. Comput. Vis. Pattern Recog.})

@String(ICCV= {Int. Conf. Comput. Vis.})

@String(TOG= {ACM Trans. Graph.})

@String(ICME = {Int. Conf. Multimedia and Expo})

@String(AAAI = {AAAI})

@String(VR   = {Vis. Res.})

@String(CVPR  = {CVPR})

@String(ICCV  = {ICCV})

@String(TOG   = {ACM TOG})

@String(ICME  =	{ICME})

@inproceedings{lipp2011interactive,
  title={Interactive modeling of city layouts using layers of procedural content},
  author={Lipp, Markus and Scherzer, Daniel and Wonka, Peter and Wimmer, Michael},
  booktitle={Computer Graphics Forum},
  volume={30},
  number={2},
  pages={345--354},
  year={2011},
  organization={Wiley Online Library}
}

@inproceedings{parish2001procedural,
  title={Procedural modeling of cities},
  author={Parish, Yoav IH and M{\"u}ller, Pascal},
  booktitle={Proceedings of the 28th annual conference on Computer graphics and interactive techniques},
  pages={301--308},
  year={2001}
}

@inproceedings{groenewegen2009procedural,
  title={Procedural City Layout Generation Based on Urban Land Use Models.},
  author={Groenewegen, Saskia and Smelik, Ruben Micha{\"e}l and de Kraker, Klaas Jan and Bidarra, Rafael},
  booktitle={Eurographics (Short Papers)},
  pages={45--48},
  year={2009}
}

@article{talton2011metropolis,
  title={Metropolis procedural modeling.},
  author={Talton, Jerry O and Lou, Yu and Lesser, Steve and Duke, Jared and Mech, Radom{\'\i}r and Koltun, Vladlen},
  journal={ACM Trans. Graph.},
  volume={30},
  number={2},
  pages={11--1},
  year={2011}
}

@article{aliaga2008interactive,
  title={Interactive example-based urban layout synthesis},
  author={Aliaga, Daniel G and Vanegas, Carlos A and Benes, Bedrich},
  journal={ACM transactions on graphics (TOG)},
  volume={27},
  number={5},
  pages={1--10},
  year={2008},
  publisher={ACM New York, NY, USA}
}

@inproceedings{vezhnevets2007interactive,
  title={Interactive image-based urban modeling},
  author={Vezhnevets, Vladimir and Konushin, Anton and Ignatenko, Alexey},
  booktitle={Proc. of PIA},
  pages={63--68},
  year={2007}
}

@article{fan2014structure,
  title={Structure completion for facade layouts.},
  author={Fan, Lubin and Musialski, Przemyslaw and Liu, Ligang and Wonka, Peter},
  journal={ACM Trans. Graph.},
  volume={33},
  number={6},
  pages={210--1},
  year={2014}
}

@inproceedings{lin2023infinicity,
  title={Infinicity: Infinite-scale city synthesis},
  author={Lin, Chieh Hubert and Lee, Hsin-Ying and Menapace, Willi and Chai, Menglei and Siarohin, Aliaksandr and Yang, Ming-Hsuan and Tulyakov, Sergey},
  booktitle={Proceedings of the IEEE/CVF international conference on computer vision},
  pages={22808--22818},
  year={2023}
}

@inproceedings{xie2024citydreamer,
  title={Citydreamer: Compositional generative model of unbounded 3d cities},
  author={Xie, Haozhe and Chen, Zhaoxi and Hong, Fangzhou and Liu, Ziwei},
  booktitle={Proceedings of the IEEE/CVF conference on computer vision and pattern recognition},
  pages={9666--9675},
  year={2024}
}

@article{deng2024citycraft,
  title={Citycraft: A real crafter for 3d city generation},
  author={Deng, Jie and Chai, Wenhao and Huang, Junsheng and Zhao, Zhonghan and Huang, Qixuan and Gao, Mingyan and Guo, Jianshu and Hao, Shengyu and Hu, Wenhao and Hwang, Jenq-Neng and others},
  journal={arXiv preprint arXiv:2406.04983},
  year={2024}
}

@InProceedings{Deng_2025_CVPR,
    author    = {Deng, Jie and Chai, Wenhao and Guo, Jianshu and Huang, Qixuan and Huang, Junsheng and Hu, Wenhao and Hao, Shengyu and Hwang, Jenq-Neng and Wang, Gaoang},
    title     = {CityGen: Infinite and Controllable City Layout Generation},
    booktitle = {Proceedings of the IEEE/CVF Conference on Computer Vision and Pattern Recognition (CVPR) Workshops},
    month     = {June},
    year      = {2025},
    pages     = {2020-2030}
}

@article{hua2025sat2city,
  title={Sat2city: 3d city generation from a single satellite image with cascaded latent diffusion},
  author={Hua, Tongyan and Jiang, Lutao and Chen, Ying-Cong and Zhao, Wufan},
  journal={arXiv preprint arXiv:2507.04403},
  year={2025}
}

@article{shen2022sgam,
  title={SGAM: Building a virtual 3d world through simultaneous generation and mapping},
  author={Shen, Yuan and Ma, Wei-Chiu and Wang, Shenlong},
  journal={Advances in Neural Information Processing Systems},
  volume={35},
  pages={22090--22102},
  year={2022}
}

@article{lee2025nuiscene,
  title={NuiScene: Exploring efficient generation of unbounded outdoor scenes},
  author={Lee, Han-Hung and Han, Qinghong and Chang, Angel X},
  journal={arXiv preprint arXiv:2503.16375},
  year={2025}
}

@article{wu2024blockfusion,
  title={Blockfusion: Expandable 3d scene generation using latent tri-plane extrapolation},
  author={Wu, Zhennan and Li, Yang and Yan, Han and Shang, Taizhang and Sun, Weixuan and Wang, Senbo and Cui, Ruikai and Liu, Weizhe and Sato, Hiroyuki and Li, Hongdong and others},
  journal={ACM Transactions on Graphics (ToG)},
  volume={43},
  number={4},
  pages={1--17},
  year={2024},
  publisher={ACM New York, NY, USA}
}

@article{yao2025cast,
  title={Cast: Component-aligned 3d scene reconstruction from an rgb image},
  author={Yao, Kaixin and Zhang, Longwen and Yan, Xinhao and Zeng, Yan and Zhang, Qixuan and Xu, Lan and Yang, Wei and Gu, Jiayuan and Yu, Jingyi},
  journal={ACM Transactions on Graphics (TOG)},
  volume={44},
  number={4},
  pages={1--19},
  year={2025},
  publisher={ACM New York, NY, USA}
}

@article{ling2025scenethesis,
  title={Scenethesis: A language and vision agentic framework for 3d scene generation},
  author={Ling, Lu and Lin, Chen-Hsuan and Lin, Tsung-Yi and Ding, Yifan and Zeng, Yu and Sheng, Yichen and Ge, Yunhao and Liu, Ming-Yu and Bera, Aniket and Li, Zhaoshuo},
  journal={arXiv preprint arXiv:2505.02836},
  year={2025}
}

@inproceedings{gu2025artiscene,
  title={ArtiScene: Language-Driven Artistic 3D Scene Generation Through Image Intermediary},
  author={Gu, Zeqi and Cui, Yin and Li, Zhaoshuo and Wei, Fangyin and Ge, Yunhao and Gu, Jinwei and Liu, Ming-Yu and Davis, Abe and Ding, Yifan},
  booktitle={Proceedings of the Computer Vision and Pattern Recognition Conference},
  pages={2891--2901},
  year={2025}
}

@article{bian2025holodeck,
  title={HOLODECK 2.0: Vision-Language-Guided 3D World Generation with Editing},
  author={Bian, Zixuan and Ren, Ruohan and Yang, Yue and Callison-Burch, Chris},
  journal={arXiv preprint arXiv:2508.05899},
  year={2025}
}

@inproceedings{xiang2025structured,
  title={Structured 3d latents for scalable and versatile 3d generation},
  author={Xiang, Jianfeng and Lv, Zelong and Xu, Sicheng and Deng, Yu and Wang, Ruicheng and Zhang, Bowen and Chen, Dong and Tong, Xin and Yang, Jiaolong},
  booktitle={Proceedings of the Computer Vision and Pattern Recognition Conference},
  pages={21469--21480},
  year={2025}
}

@article{lai2025hunyuan3d,
  title={Hunyuan3D 2.5: Towards High-Fidelity 3D Assets Generation with Ultimate Details},
  author={Lai, Zeqiang and Zhao, Yunfei and Liu, Haolin and Zhao, Zibo and Lin, Qingxiang and Shi, Huiwen and Yang, Xianghui and Yang, Mingxin and Yang, Shuhui and Feng, Yifei and others},
  journal={arXiv preprint arXiv:2506.16504},
  year={2025}
}

@article{lee2025skyfall,
  title={Skyfall-gs: Synthesizing immersive 3d urban scenes from satellite imagery},
  author={Lee, Jie-Ying and Liu, Yi-Ruei and Tsai, Shr-Ruei and Chang, Wei-Cheng and Wu, Chung-Ho and Chan, Jiewen and Zhao, Zhenjun and Lin, Chieh Hubert and Liu, Yu-Lun},
  journal={arXiv preprint arXiv:2510.15869},
  year={2025}
}

@article{achiam2023gpt,
  title={Gpt-4 technical report},
  author={Achiam, Josh and Adler, Steven and Agarwal, Sandhini and Ahmad, Lama and Akkaya, Ilge and Aleman, Florencia Leoni and Almeida, Diogo and Altenschmidt, Janko and Altman, Sam and Anadkat, Shyamal and others},
  journal={arXiv preprint arXiv:2303.08774},
  year={2023}
}

@article{floridi2020gpt,
  title={GPT-3: Its nature, scope, limits, and consequences},
  author={Floridi, Luciano and Chiriatti, Massimo},
  journal={Minds and machines},
  volume={30},
  number={4},
  pages={681--694},
  year={2020},
  publisher={Springer}
}

@article{bai2025qwen3,
  title={Qwen3-vl technical report},
  author={Bai, Shuai and Cai, Yuxuan and Chen, Ruizhe and Chen, Keqin and Chen, Xionghui and Cheng, Zesen and Deng, Lianghao and Ding, Wei and Gao, Chang and Ge, Chunjiang and others},
  journal={arXiv preprint arXiv:2511.21631},
  year={2025}
}

@article{fu2025mano,
  title={Mano Technical Report},
  author={Fu, Tianyu and Su, Anyang and Zhao, Chenxu and Wang, Hanning and Wu, Minghui and Yu, Zhe and Hu, Fei and Shi, Mingjia and Dong, Wei and Wang, Jiayao and others},
  journal={arXiv preprint arXiv:2509.17336},
  year={2025}
}

@article{zhao2025resilient,
  title={Resilient odometry via hierarchical adaptation},
  author={Zhao, Shibo and Zhou, Sifan and Zhang, Yuchen and Zhang, Ji and Wang, Chen and Wang, Wenshan and Scherer, Sebastian},
  journal={Science Robotics},
  volume={10},
  number={109},
  pages={eadv1818},
  year={2025},
  publisher={American Association for the Advancement of Science}
}

@article{zhang2024codeagent,
  title={Codeagent: Enhancing code generation with tool-integrated agent systems for real-world repo-level coding challenges},
  author={Zhang, Kechi and Li, Jia and Li, Ge and Shi, Xianjie and Jin, Zhi},
  journal={arXiv preprint arXiv:2401.07339},
  year={2024}
}

@article{qian2023chatdev,
  title={Chatdev: Communicative agents for software development},
  author={Qian, Chen and Liu, Wei and Liu, Hongzhang and Chen, Nuo and Dang, Yufan and Li, Jiahao and Yang, Cheng and Chen, Weize and Su, Yusheng and Cong, Xin and others},
  journal={arXiv preprint arXiv:2307.07924},
  year={2023}
}

@article{yang2024swe,
  title={Swe-agent: Agent-computer interfaces enable automated software engineering},
  author={Yang, John and Jimenez, Carlos E and Wettig, Alexander and Lieret, Kilian and Yao, Shunyu and Narasimhan, Karthik and Press, Ofir},
  journal={Advances in Neural Information Processing Systems},
  volume={37},
  pages={50528--50652},
  year={2024}
}

@article{kim2024image,
  title={An image grid can be worth a video: Zero-shot video question answering using a vlm},
  author={Kim, Wonkyun and Choi, Changin and Lee, Wonseok and Rhee, Wonjong},
  journal={IEEE Access},
  year={2024},
  publisher={IEEE}
}

@article{feng2023layoutgpt,
  title={Layoutgpt: Compositional visual planning and generation with large language models},
  author={Feng, Weixi and Zhu, Wanrong and Fu, Tsu-jui and Jampani, Varun and Akula, Arjun and He, Xuehai and Basu, Sugato and Wang, Xin Eric and Wang, William Yang},
  journal={Advances in Neural Information Processing Systems},
  volume={36},
  pages={18225--18250},
  year={2023}
}

@inproceedings{sun2025layoutvlm,
  title={Layoutvlm: Differentiable optimization of 3d layout via vision-language models},
  author={Sun, Fan-Yun and Liu, Weiyu and Gu, Siyi and Lim, Dylan and Bhat, Goutam and Tombari, Federico and Li, Manling and Haber, Nick and Wu, Jiajun},
  booktitle={Proceedings of the Computer Vision and Pattern Recognition Conference},
  pages={29469--29478},
  year={2025}
}

@inproceedings{ccelen2024design,
  title={I-design: Personalized llm interior designer},
  author={{\c{C}}elen, Ata and Han, Guo and Schindler, Konrad and Van Gool, Luc and Armeni, Iro and Obukhov, Anton and Wang, Xi},
  booktitle={European Conference on Computer Vision},
  pages={217--234},
  year={2024},
  organization={Springer}
}

@inproceedings{yang2024holodeck,
  title={Holodeck: Language guided generation of 3d embodied ai environments},
  author={Yang, Yue and Sun, Fan-Yun and Weihs, Luca and VanderBilt, Eli and Herrasti, Alvaro and Han, Winson and Wu, Jiajun and Haber, Nick and Krishna, Ranjay and Liu, Lingjie and others},
  booktitle={Proceedings of the IEEE/CVF Conference on Computer Vision and Pattern Recognition},
  pages={16227--16237},
  year={2024}
}

@article{pan2025metaspatial,
  title={Metaspatial: Reinforcing 3d spatial reasoning in vlms for the metaverse},
  author={Pan, Zhenyu and Liu, Han},
  journal={arXiv preprint arXiv:2503.18470},
  year={2025}
}

@article{paschalidou2021atiss,
  title={Atiss: Autoregressive transformers for indoor scene synthesis},
  author={Paschalidou, Despoina and Kar, Amlan and Shugrina, Maria and Kreis, Karsten and Geiger, Andreas and Fidler, Sanja},
  journal={Advances in Neural Information Processing Systems},
  volume={34},
  pages={12013--12026},
  year={2021}
}

@inproceedings{tang2024diffuscene,
  title={Diffuscene: Denoising diffusion models for generative indoor scene synthesis},
  author={Tang, Jiapeng and Nie, Yinyu and Markhasin, Lev and Dai, Angela and Thies, Justus and Nie{\ss}ner, Matthias},
  booktitle={Proceedings of the IEEE/CVF conference on computer vision and pattern recognition},
  pages={20507--20518},
  year={2024}
}

@inproceedings{yang2025mmgdreamer,
  title={Mmgdreamer: Mixed-modality graph for geometry-controllable 3d indoor scene generation},
  author={Yang, Zhifei and Lu, Keyang and Zhang, Chao and Qi, Jiaxing and Jiang, Hanqi and Ma, Ruifei and Yin, Shenglin and Xu, Yifan and Xing, Mingzhe and Xiao, Zhen and others},
  booktitle={Proceedings of the AAAI Conference on Artificial Intelligence},
  volume={39},
  number={9},
  pages={9391--9399},
  year={2025}
}

@inproceedings{yu2025wonderworld,
  title={Wonderworld: Interactive 3d scene generation from a single image},
  author={Yu, Hong-Xing and Duan, Haoyi and Herrmann, Charles and Freeman, William T and Wu, Jiajun},
  booktitle={Proceedings of the Computer Vision and Pattern Recognition Conference},
  pages={5916--5926},
  year={2025}
}

@inproceedings{zhou2025scenex,
  title={Scenex: Procedural controllable large-scale scene generation},
  author={Zhou, Mengqi and Wang, Yuxi and Hou, Jun and Zhang, Shougao and Li, Yiwei and Luo, Chuanchen and Peng, Junran and Zhang, Zhaoxiang},
  booktitle={Proceedings of the AAAI Conference on Artificial Intelligence},
  volume={39},
  number={10},
  pages={10806--10814},
  year={2025}
}

@InProceedings{Chai_2023_CVPR,
    author    = {Chai, Lucy and Tucker, Richard and Li, Zhengqi and Isola, Phillip and Snavely, Noah},
    title     = {Persistent Nature: A Generative Model of Unbounded 3D Worlds},
    booktitle = {Proceedings of the IEEE/CVF Conference on Computer Vision and Pattern Recognition (CVPR)},
    month     = {June},
    year      = {2023},
    pages     = {20863-20874}
}

@ARTICLE{10269790,
  author={Chen, Zhaoxi and Wang, Guangcong and Liu, Ziwei},
  journal={IEEE Transactions on Pattern Analysis and Machine Intelligence}, 
  title={SceneDreamer: Unbounded 3D Scene Generation From 2D Image Collections}, 
  year={2023},
  volume={45},
  number={12},
  pages={15562-15576},
  doi={10.1109/TPAMI.2023.3321857}}

@InProceedings{Yang_2024_CVPR,
    author    = {Yang, Yandan and Jia, Baoxiong and Zhi, Peiyuan and Huang, Siyuan},
    title     = {PhyScene: Physically Interactable 3D Scene Synthesis for Embodied AI},
    booktitle = {Proceedings of the IEEE/CVF Conference on Computer Vision and Pattern Recognition (CVPR)},
    month     = {June},
    year      = {2024},
    pages     = {16262-16272}
}

@article{ghorbanian2019procedural,
  title={Procedural modeling as a practical technique for 3D assessment in urban design via CityEngine},
  author={Ghorbanian, M and Shariatpour, F and others},
  journal={Int. J. Architect. Eng. Urban Plan},
  volume={29},
  number={2},
  pages={255--267},
  year={2019}
}

@incollection{kelly2021cityengine,
  title={CityEngine: An introduction to rule-based modeling},
  author={Kelly, Tom},
  booktitle={Urban informatics},
  pages={637--662},
  year={2021},
  publisher={Springer}
}

@article{niu2025controllable,
  title={Controllable Generation of Large-Scale 3D Urban Layouts with Semantic and Structural Guidance},
  author={Niu, Mengyuan and Zhuo, Xinxin and Wang, Ruizhe and Huang, Yuyue and Yang, Junyan and Wang, Qiao},
  journal={arXiv preprint arXiv:2509.23804},
  year={2025}
}

@article{touvron2023llama,
  title={Llama: Open and efficient foundation language models},
  author={Touvron, Hugo and Lavril, Thibaut and Izacard, Gautier and Martinet, Xavier and Lachaux, Marie-Anne and Lacroix, Timoth{\'e}e and Rozi{\`e}re, Baptiste and Goyal, Naman and Hambro, Eric and Azhar, Faisal and others},
  journal={arXiv preprint arXiv:2302.13971},
  year={2023}
}

@article{anil2023palm,
  title={Palm 2 technical report},
  author={Anil, Rohan and Dai, Andrew M and Firat, Orhan and Johnson, Melvin and Lepikhin, Dmitry and Passos, Alexandre and Shakeri, Siamak and Taropa, Emanuel and Bailey, Paige and Chen, Zhifeng and others},
  journal={arXiv preprint arXiv:2305.10403},
  year={2023}
}

@article{liu2023visual,
  title={Visual instruction tuning},
  author={Liu, Haotian and Li, Chunyuan and Wu, Qingyang and Lee, Yong Jae},
  journal={Advances in neural information processing systems},
  volume={36},
  pages={34892--34916},
  year={2023}
}

@article{li2023llava,
  title={Llava-med: Training a large language-and-vision assistant for biomedicine in one day},
  author={Li, Chunyuan and Wong, Cliff and Zhang, Sheng and Usuyama, Naoto and Liu, Haotian and Yang, Jianwei and Naumann, Tristan and Poon, Hoifung and Gao, Jianfeng},
  journal={Advances in Neural Information Processing Systems},
  volume={36},
  pages={28541--28564},
  year={2023}
}

@inproceedings{yin2025spatial,
  title={Spatial mental modeling from limited views},
  author={Yin, Baiqiao and Wang, Qineng and Zhang, Pingyue and Zhang, Jianshu and Wang, Kangrui and Wang, Zihan and Zhang, Jieyu and Chandrasegaran, Keshigeyan and Liu, Han and Krishna, Ranjay and others},
  booktitle={Structural Priors for Vision Workshop at ICCV'25},
  year={2025}
}

@article{pan2025metafind,
  title={MetaFind: Scene-Aware 3D Asset Retrieval for Coherent Metaverse Scene Generation},
  author={Pan, Zhenyu and Lu, Yucheng and Liu, Han},
  journal={arXiv preprint arXiv:2510.04057},
  year={2025}
}

@inproceedings{vincur2017vr,
  title={VR City: Software analysis in virtual reality environment},
  author={Vincur, Juraj and Navrat, Pavol and Polasek, Ivan},
  booktitle={2017 IEEE international conference on software quality, reliability and security companion (QRS-C)},
  pages={509--516},
  year={2017},
  organization={IEEE}
}

@inproceedings{tan2016evolution,
  title={The evolution of city gaming},
  author={Tan, Ekim},
  booktitle={Complexity, Cognition, Urban Planning and Design: Post-Proceedings of the 2nd Delft International Conference},
  pages={271--292},
  year={2016},
  organization={Springer}
}

@article{shan2021research,
  title={Research on 3D urban landscape design and evaluation based on geographic information system},
  author={Shan, Pengyu and Sun, Wan},
  journal={Environmental Earth Sciences},
  volume={80},
  number={17},
  pages={597},
  year={2021},
  publisher={Springer}
}

@article{schrotter2020digital,
  title={The digital twin of the city of Zurich for urban planning},
  author={Schrotter, Gerhard and H{\"u}rzeler, Christian},
  journal={PFG--Journal of Photogrammetry, Remote Sensing and Geoinformation Science},
  volume={88},
  number={1},
  pages={99--112},
  year={2020},
  publisher={Springer}
}

@inproceedings{lin2024evaluating,
  title={Evaluating text-to-visual generation with image-to-text generation},
  author={Lin, Zhiqiu and Pathak, Deepak and Li, Baiqi and Li, Jiayao and Xia, Xide and Neubig, Graham and Zhang, Pengchuan and Ramanan, Deva},
  booktitle={European Conference on Computer Vision},
  pages={366--384},
  year={2024},
  organization={Springer}
}

@article{seo2025paper2code,
  title={Paper2code: Automating code generation from scientific papers in machine learning},
  author={Seo, Minju and Baek, Jinheon and Lee, Seongyun and Hwang, Sung Ju},
  journal={arXiv preprint arXiv:2504.17192},
  year={2025}
}

@article{yue2025foam,
  title={Foam-Agent: Towards Automated Intelligent CFD Workflows},
  author={Yue, Ling and Somasekharan, Nithin and Cao, Yadi and Pan, Shaowu},
  journal={arXiv preprint arXiv:2505.04997},
  year={2025}
}

@article{su2025openthinkimg,
  title={Openthinkimg: Learning to think with images via visual tool reinforcement learning},
  author={Su, Zhaochen and Li, Linjie and Song, Mingyang and Hao, Yunzhuo and Yang, Zhengyuan and Zhang, Jun and Chen, Guanjie and Gu, Jiawei and Li, Juntao and Qu, Xiaoye and others},
  journal={arXiv preprint arXiv:2505.08617},
  year={2025}
}

@article{huang2025visualtoolagent,
  title={Visualtoolagent (vista): A reinforcement learning framework for visual tool selection},
  author={Huang, Zeyi and Ji, Yuyang and Rajan, Anirudh Sundara and Cai, Zefan and Xiao, Wen and Wang, Haohan and Hu, Junjie and Lee, Yong Jae},
  journal={arXiv preprint arXiv:2505.20289},
  year={2025}
}

@article{wu2025vtool,
  title={VTool-R1: VLMs Learn to Think with Images via Reinforcement Learning on Multimodal Tool Use},
  author={Wu, Mingyuan and Yang, Jingcheng and Jiang, Jize and Li, Meitang and Yan, Kaizhuo and Yu, Hanchao and Zhang, Minjia and Zhai, Chengxiang and Nahrstedt, Klara},
  journal={arXiv preprint arXiv:2505.19255},
  year={2025}
}

@inproceedings{koh2025c2,
  title={c2: Scalable auto-feedback for LLM-based chart generation},
  author={Koh, Woosung and Yoon, Janghan and Lee, MinHyung and Song, Youngjin and Cho, Jaegwan and Kang, Jaehyun and Kim, Taehyeon and Yun, Se-Young and Yu, Youngjae and Lee, Bongshin},
  booktitle={Proceedings of the 2025 Conference of the Nations of the Americas Chapter of the Association for Computational Linguistics: Human Language Technologies (Volume 1: Long Papers)},
  pages={4525--4566},
  year={2025}
}

@inproceedings{reimers2019sentence,
  title={Sentence-bert: Sentence embeddings using siamese bert-networks},
  author={Reimers, Nils and Gurevych, Iryna},
  booktitle={Proceedings of the 2019 conference on empirical methods in natural language processing and the 9th international joint conference on natural language processing (EMNLP-IJCNLP)},
  pages={3982--3992},
  year={2019}
}

@InProceedings{Engstler_2025_ICCV,
    author    = {Engstler, Paul and Shtedritski, Aleksandar and Laina, Iro and Rupprecht, Christian and Vedaldi, Andrea},
    title     = {SynCity: Training-Free Generation of 3D Worlds},
    booktitle = {Proceedings of the IEEE/CVF International Conference on Computer Vision (ICCV)},
    month     = {October},
    year      = {2025},
    pages     = {27585-27595}
}

@article{singh2025openai,
  title={Openai gpt-5 system card},
  author={Singh, Aaditya and Fry, Adam and Perelman, Adam and Tart, Adam and Ganesh, Adi and El-Kishky, Ahmed and McLaughlin, Aidan and Low, Aiden and Ostrow, AJ and Ananthram, Akhila and others},
  journal={arXiv preprint arXiv:2601.03267},
  year={2025}
}

@INPROCEEDINGS{11208990,
  author={Wang, Xiaoyan and Li, Zeju and Xu, Yifan and Qi, Jiaxing and Yang, Zhifei and Ma, Ruifei and Liu, Xiangde and Zhang, Chao},
  booktitle={2025 IEEE International Conference on Multimedia and Expo (ICME)}, 
  title={Spatial 3D-LLM : Exploring Spatial Awareness in 3D Vision-Language Models}, 
  year={2025},
  volume={},
  number={},
  pages={1-6},
  doi={10.1109/ICME59968.2025.11208990}}

@InProceedings{Fang_2023_ICCV,
    author    = {Fang, Han and Yang, Zhifei and Wei, Yuhan and Zang, Xianghao and Ban, Chao and Feng, Zerun and He, Zhongjiang and Li, Yongxiang and Sun, Hao},
    title     = {Alignment and Generation Adapter for Efficient Video-Text Understanding},
    booktitle = {Proceedings of the IEEE/CVF International Conference on Computer Vision (ICCV) Workshops},
    month     = {October},
    year      = {2023},
    pages     = {2791-2797}
}

@inproceedings{10.1145/3581783.3611756, author = {Fang, Han and Yang, Zhifei and Zang, Xianghao and Ban, Chao and He, Zhongjiang and Sun, Hao and Zhou, Lanxiang}, title = {Mask to Reconstruct: Cooperative Semantics Completion for Video-text Retrieval}, year = {2023}, isbn = {9798400701085}, publisher = {Association for Computing Machinery}, address = {New York, NY, USA}, url = {https://doi.org/10.1145/3581783.3611756}, doi = {10.1145/3581783.3611756}, booktitle = {Proceedings of the 31st ACM International Conference on Multimedia}, pages = {3847–3856}, numpages = {10}, location = {Ottawa ON, Canada}, series = {MM '23} }

@article{yang2025vidtext,
  title={Vidtext: Towards comprehensive evaluation for video text understanding},
  author={Yang, Zhoufaran and Shu, Yan and Wang, Jing and Yang, Zhifei and Zhang, Yan and Li, Yu and Lu, Keyang and Zeng, Gangyan and Liu, Shaohui and Zhou, Yu and others},
  journal={arXiv preprint arXiv:2505.22810},
  year={2025}
}

@inproceedings{ma2025ctsg,
  title={CTSG: Integrating Context and Way Topology Into Scene Graph for Zero-shot Navigation},
  author={Ma, Ruifei and Xu, Yifan and Li, Yuze and Fang, Yanping and Yang, Zhifei and Qi, Jiaxing and Zhao, Xinyu and Zhang, Chao},
  booktitle={2025 IEEE/RSJ International Conference on Intelligent Robots and Systems (IROS)},
  pages={18934--18941},
  year={2025},
  organization={IEEE}
}
}
\clearpage
\maketitlesupplementary
\setcounter{section}{0}
\renewcommand{\thesection}{\Alph{section}}
\renewcommand{\thesubsection}{\Alph{section}.\arabic{subsection}}
\section{Additional Results}
\subsection{Qualitative Comparison}
\label{Supplementary_Qualitative_Comparison}
\begin{figure*}
    \centering
    \includegraphics[width=0.9\linewidth]{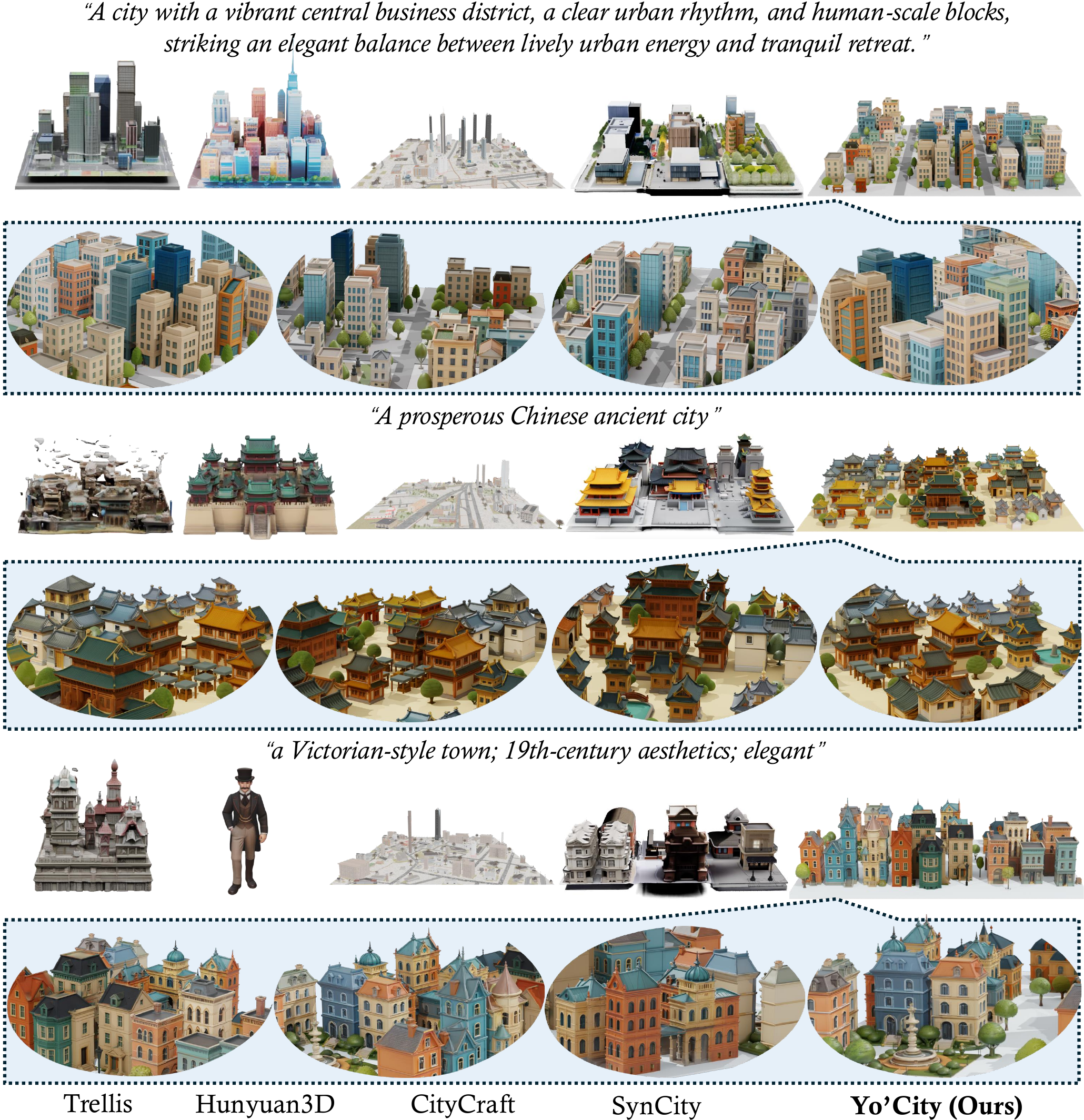}
    \caption{\textbf{Additional qualitative comparison between our method and the baselines}. These three cases correspond to the three major city instruction types represented in our dataset. We highlight the zoom-in views of Yo’City results from four different perspectives.}
    \label{fig:supplementary_material_case}
    \vspace{-4mm}
\end{figure*}

We present three qualitative visual comparisons in Fig.~\ref{fig:supplementary_material_case} and further provide rendered Yo’City results from four different viewpoints. The city instructions correspond to three different architectural styles, namely a modern city, a Chinese ancient city, and a 19th-century European town. Our results show clear superiority in fidelity and have more details. For example, in the modern city case, Yo’City produces a more reasonable layout with well-coordinated buildings and efficient spatial utilization, while baseline methods often lead to disorganized structures or inconsistent urban patterns. Similarly, in the Victorian-style town, Yo’City maintains coherent façade details and roof shapes, demonstrating better fidelity and style control. These results validate the effectiveness of our design. 
\subsection{Visualization of Expansion}
\begin{figure*}
    \centering
    \includegraphics[width=\linewidth]{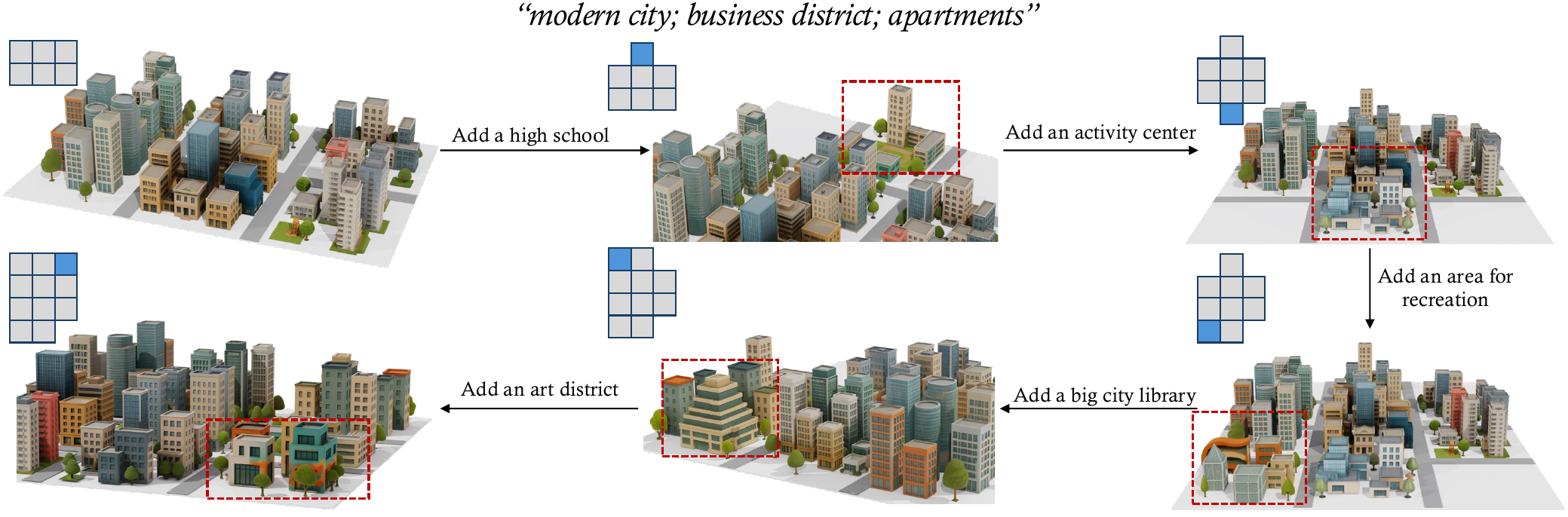}
    \caption{\textbf{Additional visualization of expansion}. The first row presents the city’s global instruction. The leftmost city shows the initial generation result, followed by four successive expansion iterations. In the top-left corner, a BEV thumbnail depicts the city layout, with blue regions indicating newly expanded grids, while red boxes in the rendered images highlight their appearances.
    } 
    \label{fig:supplementary_ablation_expansion}
    \vspace{-3mm}
\end{figure*}

Fig.~\ref{fig:supplementary_ablation_expansion} shows an example of city expansion. Through four successive iterations, Yo'City successfully preserves the integrity of the original global instructions while progressively refining the city's design. Each expansion introduces new elements in a coherent and visually consistent manner, aligning with the overarching goals of urban vitality and quality. Our self-critic mechanism, featuring distance- and semantics-aware optimization, ensures that each step identifies the optimal location for a new grid based on the existing scene. This approach enables a careful consideration of the spatial relationships and surrounding contexts, ensuring that each expansion fits naturally and harmoniously without any abrupt transitions. Empowered by it, our model can continuously reason about existing results and extrapolate from them, achieving unbounded generation.

\subsection{Expansion Mechanism Comparison}
We compare our relationship-guided expansion mechanism with random placement and a semantic-only strategy. We report the Coefficient of Variance (CV) of the VQAScore~\cite{lin2024evaluating} to evaluate generation quality and stability. As shown in Tab.~\ref{tab:cv_expansion}, our method achieves the lowest CV, indicating its effectiveness and greater robustness. 

\begin{table}[t]
\centering
\caption{\textbf{Stability Comparison Across Expansion Strategies.} We report the coefficient of variance (CV) of VQAScore.}
\label{tab:cv_expansion}
\setlength{\tabcolsep}{6pt}
\renewcommand{\arraystretch}{0.9}
\begin{tabular}{lc}
\toprule
Expansion Strategy & CV $\downarrow$ \\
\midrule
Random & 8.76\% \\
Semantics-Only & 5.16\% \\
\rowcolor{myblue!18} Ours & \textbf{3.34\%} \\
\bottomrule
\end{tabular}
\end{table}

\begin{figure}[ht]
\centering
\begin{tcolorbox}[
    colback=gray!5,
    colframe=gray!60,
    title=\textbf{Prompt for Generating Your City Instructions},
    fonttitle=\bfseries,
]
\small
Generate 5 diverse city-design instructions. Each instruction should focus primarily on intrinsic city characteristics—such as functional zoning, architectural typologies, structural composition—rather than broad geographical environments.
You may freely mix description styles (short sentences, extended sentences, keyword lists), incorporating stylistic attributes (eg. "modern", "ancient"), city scales (e.g., “Size 2×3”), aesthetic tendencies, structural patterns into the instructions.

\medskip
\textbf{Examples:}

\textit{Short Sentence Example:}  
``A dynamic business city.''  

\textit{Long Sentence Example:}  
``A modern city with a stylized central business district, white high-rise residential buildings, 
and a convenient activity center.''  

\textit{Keywords-based Example:}  
``modern city; entertainment hubs; high-rise buildings; apartments''

\medskip
\textbf{Output Format.}
\begin{verbatim}
1. ...
2. ...
3. ...
4. ...
5. ...
\end{verbatim}
\end{tcolorbox}
\vspace{-4mm}
\caption{A template for generating various city instructions.}
\vspace{-3mm}
\label{dataset_curation_prompt}
\end{figure}

\section{Dataset Curation Details}
\label{dataset_curation}
We construct a text dataset containing multiple types of descriptions to evaluate different methods. Among them, 30\% are written by humans and used as few-shot examples for GPT-4o~\cite{achiam2023gpt} to generate the remaining 70\%. To enable comprehensive evaluation, this dataset comprises multiple forms of text, as follows:

\noindent\textbf{Short Sentence}. This refers to a concise sentence expressing the generation requirement, such as “\textit{a modern city}” or “\textit{a vibrant business city}.” To enrich its diversity, some descriptions specify the scene size, such as “\textit{Size 2 × 3, a cozy city}.” Others include stylistic references, such as “\textit{a town in the style of Disneyland}” or “\textit{a Beijing-like big city}.”

\noindent\textbf{Long Sentence}. These typically include an overall description of the scene along with specific detailed requirements, and are used to assess whether the model can process complicated inputs and capture users' personalized intentions. For example: “\textit{A modern city featuring skyscrapers and a bustling entertainment district, with diverse architectural styles and a realistic urban layout},” or “\textit{A prosperous ancient Chinese town with tiled-roof houses, lively markets, and ornate gates}.” 

We fully consider the maximum input length limitations of baseline methods, and therefore control the sentence length in our dataset. However, in practice, our method can accommodate much longer inputs, including paragraph-level descriptions.

\noindent\textbf{Keywords-based Description}. We also include keyword-style inputs in the dataset to better adapt to diverse user input habits, such as “\textit{modern city; entertainment hubs; high-rise buildings; convenient life}.”

Here, we provide a prompt in Fig.\ref{dataset_curation_prompt} that can be used to generate diverse city instructions.

\section{Implementation Details}
\label{implementation}
\noindent\textbf{Hardware Setup}. All experiments in this paper are conducted on a server running Ubuntu 22.04, equipped with an Intel(R) Xeon(R) CPU E5-2699 v4 @ 2.20GHz,  and NVIDIA A6000 GPUs with 48GB of memory.

\noindent\textbf{Prompt Templates}. In our experiments, we used GPT-4o as the LLM and GPT-Image-1 as the model for image generation and editing, both accessed via the official APIs. ~\ref{Global_Planner}, ~\ref{Local_Designer}, ~\ref{Image_Generator_1}, ~\ref{Image_Generator_2}, ~\ref{Image_Evaluator}, ~\ref{Relationship_Constraints} are the prompts we use for different agents. The black text represents the system prompt, and the blue part represents the input that the agent needs to receive.

\noindent\textbf{Image Generator}. The "produce-refine-evaluation" loop for the 3D Generator is executed up to three iterations. The Image Evaluator assigns a score from 0 to 10, with scores not less than 6 considered acceptable. If the score is below 6, we prompt VLM to rewrite the generation instruction based on the current negative feedback. The loop terminates once an acceptable score is reached. 

\noindent\textbf{Seam Artifacts Mitigation}. As illustrated in the pipeline figure in the main paper, background content and tile-related structures are removed during the refinement stage of image generation. This ensures that each generated grid contains only foreground objects, thereby preventing seam artifacts (e.g., visible grid boundaries) during stitching.

\noindent\textbf{Cross-grid Consistency}. To avoid substantial computational overhead, we don't enforce explicit cross-grid interaction in 3D generation. However, during image generation, our evaluator checks alignment between each generated image and its description from Local Designer and triggers regeneration upon detected deviations. This mechanism indirectly promotes cross-grid consistency by ensuring adherence to mutually consistent semantic specifications, effectively mitigating severe cross-grid inconsistencies.

\noindent\textbf{Post Processing}. We follow real-world urban planning strategies by first establishing specific functional districts and then connecting them through road networks. After obtaining the 3D models for each grid, we first scale them to ensure consistent proportions. Next, utilizing Blender's API, Yo'City integrates the ground, roads, and other elements. The default color for the roads in the modern city is set to [0.15, 0.15, 0.15, 1.0], which corresponds to a dark gray. The ground color is defined as [0.50, 0.50, 0.50, 1.0], representing a medium gray. Both the road and ground materials have a roughness value of 0.9. These parameters can also be customized, allowing for adjustments to the road and ground colors to better align with the desired style. Additionally, Yo'City enables users to specify which roads should be connected, supporting diverse road networks.

\noindent\textbf{Expansion Module}. 
To perform semantic regularization, we use a Sentence-Bert~\cite{reimers2019sentence} model to encode the grid descriptions and compute the cosine similarity between their [CLS] embeddings. In the optimization process, we assign weights of 1, 0.5, 0.1, 0, and -1 to the five types of spatial relationships \{near, relatively near, slightly near, no special constraint, far\}, respectively. And the regularization parameter $\lambda$ is set to 1. 
\section{Evaluation Details}
\label{evaluation}
\begin{figure*}[t]
\centering
\begin{tcolorbox}[
    colback=gray!5,
    colframe=gray!60,
    title=\textbf{Criteria to Evaluate Visual Quality},
    fonttitle=\bfseries,
    width=\textwidth,   
    boxrule=0.5pt
]
\small
\textbf{Role Definition}: 

You are an expert evaluator in 3D urban scene generation, with deep expertise in 3D generation, AIGC, and city-scale simulation.

Your task is to objectively compare two rendered images of 3D-generated city scenes and determine which method performs better overall. Both images depict city environments rendered from a frontal angle of approximately 15°. If parts of the scene are partially visible, you should infer the full structure based on visible cues to assess the entire city layout.

The evaluation should be based solely on the visual quality of the rendered 3D GLB outputs, with no further post-processing assumed. 

- Record the first image as A, and the second image as B.

- For every request you receive, reason carefully about the specified dimension, then answer with a single letter: either A (if image A is superior) or B (if image B is superior). Never output any other text beyond that single letter.

\medskip
\textbf{\textit{Geometric Fidelity}}: Evaluate only geometric fidelity. Criteria:

Which result has cleaner, more complete building shapes?

Which result has fewer distortions, floating parts, holes, or irregular ground transitions?

Which scene demonstrates more stable, natural, and well-formed geometry?

\medskip
\textbf{\textit{Texture Clarity}}: Evaluate only texture clarity. Criteria:

- Which one has sharpe and clearer textures?

- Which one shows more structural details of buildings (blurriness is unacceptable)?

- Under the premise of non-exaggerated appearance/texture, which cityscape better represents high-fidelity appearance?

\medskip
\textbf{\textit{Layout Coherence}}: Evaluate only layout coherence. Criteria: 

- Which result shows a more logical and realistic spatial arrangement of buildings and roads?

- Which one exhibits a clearer city structure or hierarchical organization?

- Which feels more like a coherent, reasonably-distributed and well-planned city?

\medskip
\textbf{\textit{Scene Coverage}}: Evaluate only scene coverage. Criteria: 

- Which city covers a larger or more complete area?

- Which has more buildings and a better sense of an extended city environment?

- Which gives a stronger impression of a full modern city?

\medskip
\textbf{\textit{Overall Realism}}:  Evaluate only overall realism. Criteria:

- When evaluating realism, focus on visual form and shape and ignore rendering effects. Realism refers to how well the building heights, architectural appearances, spacing between buildings, and overall scene layout align with real-world urban environments.

- Which result has more reasonable and plausible building heights and forms? (Avoiding overly exaggerated shapes.)

- Which result conveys a more natural and realistic overall urban atmosphere?

- Which result suggests a more complete and realistic living environment with multiple functional zones?

\medskip
\textbf{Input:}
\begin{verbatim}
1. City Instruction:
2. Image A
3. Image B
\end{verbatim}
\end{tcolorbox}
\caption{\textbf{Evaluation criteria for visual quality}, focusing on Geometric Fidelity, Texture Clarity, Layout Coherence, Scene Coverage, and Overall Realism. The prompt provides detailed assessment standards. }
\vspace{-3.mm}
\label{Evalution_Prompt}
\end{figure*}

\noindent\textbf{Semantic Consistency}. We utilize VQAScore~\cite{lin2024evaluating} to assess the semantic consistency between the generated city and the input text. Specifically, it leverages the CLIP-FlanT5 model to compute an alignment score based on the textual requirement (\textit{city instruction + “with balanced proportions and realistic, non-exaggerated forms.”}) and the corresponding rendered city image. This metric not only evaluates how accurately the generated city reflects the provided preferences, but also tests the reasonableness and realism of the output, which is essential for realistic city scene generation. To ensure a fair comparison, all images are rendered from identical viewpoints.

\noindent\textbf{Visual Quality}. To assess the visual and perceptual quality  of the generated results, we introduce five dimensions and perform pairwise evaluations, involving both VLM and human judges. For the VLM judge, we select GPT-5, which excels at understanding 3D spatial relationships and performing multimodal reasoning, enabling it to deliver a more thorough and precise evaluation. For the human judges, we recruit 10 volunteers from diverse professional backgrounds. In the evaluation process, we conduct pairwise assessments for each dimension independently. To minimize randomness, each comparison is repeated twice. We use the win rates of different methods as the quantitative metric. The specific criteria can be found in Fig.~\ref{Evalution_Prompt}. 

\noindent\textbf{Grid-level Alignment Score}. To evaluate the consistency between each grid and the global instruction, the Alignment Score is calculated using the VQAScore. This score is derived by comparing the grid image with the query, “\textit{Does this figure show a reasonable grid of \{city instruction\}?}", evaluating the model's ability to maintain consistent city-related semantics across different regions.

\noindent\textbf{Grid-level Aesthetic Score}. To complement assessments provided by GPT-5 and human judges, we introduce an Aesthetic Score specifically designed to quantify the local aesthetic quality of the generated city. This score is derived from a SigLIP-based aesthetic predictor that evaluates image visual quality on a scale of 1 to 10, with a score of 5.5+ regarded as great.

\section{More Analysis}
\subsection{Computational Efficiency}
\begingroup

We compare Yo'City with CityCraft~\cite{deng2024citycraft} and SynCity~\cite{Engstler_2025_ICCV} in terms of generation time and GPU memory consumption. 
The generation time is measured as the average time required to produce a city of typical size. 
Following the official implementations, all three methods access GPT-4o via the API, and we replace the image and 3D generation APIs in Yo'City with FLUX.2-Klein-9B\hypertarget{fn:flux_ref}{\hyperlink{fn:flux_note}{\textsuperscript{$\ddagger$}}} and Trellis~\cite{xiang2025structured} for fair memory comparison.

\footnotetext[1]{\hypertarget{fn:flux_note}{\hyperlink{fn:flux_ref}{\textsuperscript{$\ddagger$}}}\ FLUX.2-Klein-9B model from Black Forest Labs: \url{https://huggingface.co/black-forest-labs/FLUX.2-klein-9B}}

\endgroup

\begin{table}[t]
\centering
\setlength{\tabcolsep}{4pt}
\renewcommand{\arraystretch}{0.9}
\caption{\textbf{Runtime, GPU memory, and VQAScore comparison.}}
\label{tab:runtime_analysis}
\begin{tabular}{lccc}
\toprule
Method & Runtime & Memory & VQAScore \\
\midrule
CityCraft~\cite{deng2024citycraft} & 21.17 min & 4 GB  & 0.5639 \\
SynCity~\cite{Engstler_2025_ICCV}   & 62.47 min & 40 GB & 0.6975 \\
\rowcolor{myblue!18} Yo'City & 24.99 min & 24 GB & \textbf{0.7097} \\
\bottomrule
\end{tabular}
\end{table}
\begin{figure}
    \centering
    \includegraphics[width=\linewidth]{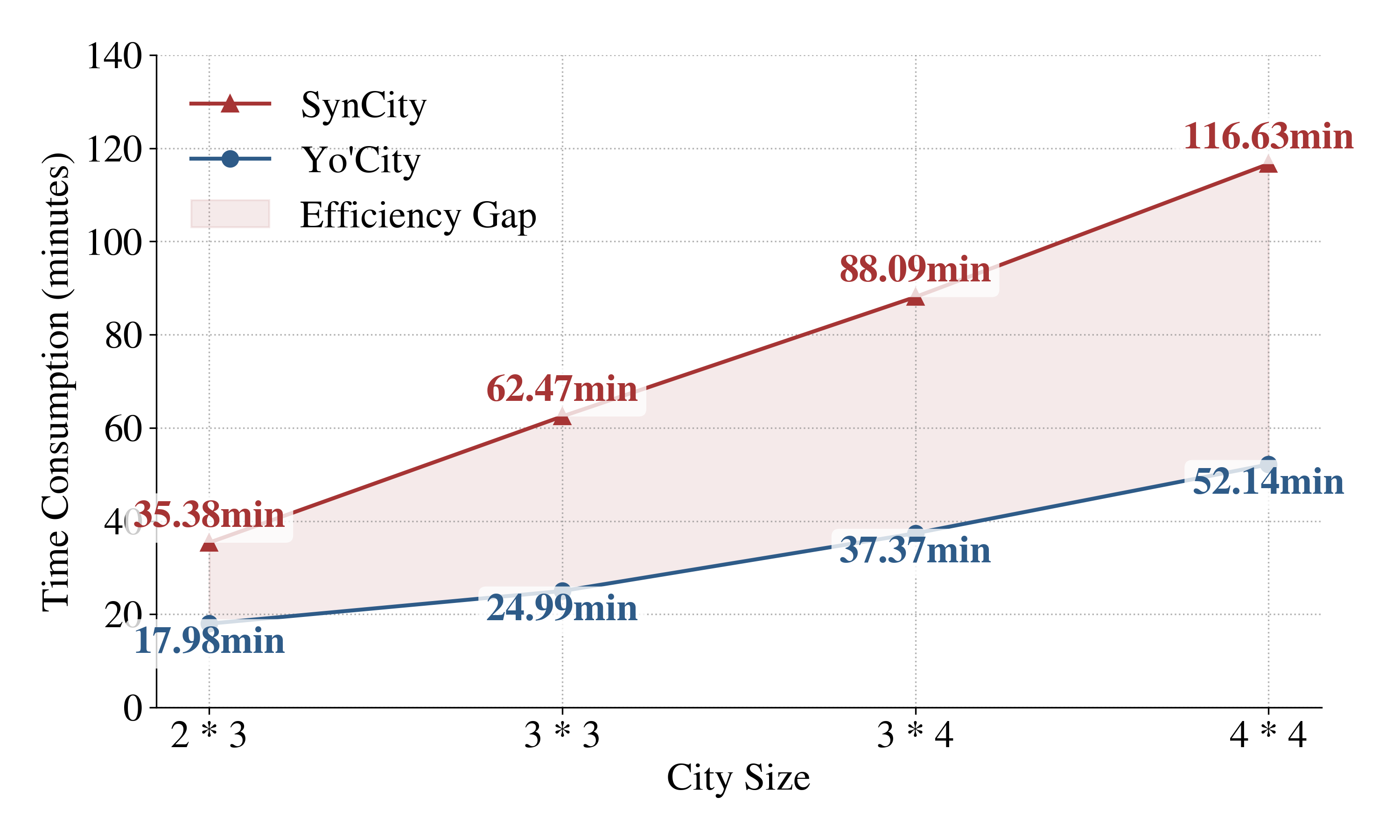}
    \caption{\textbf{Comparison of Time Consumption between SynCity and Yo’City across different city sizes}. The results demonstrate that Yo’City consistently exhibits better efficiency than SynCity.}
    \label{fig:efficiency_figure}
    \vspace{-4mm}
\end{figure}
\noindent\textbf{Detailed Discussion between Yo'City and SynCity}.
Since Yo'City is non-autoregressive, different city areas can be generated in parallel, which is difficult for SynCity. Moreover, Yo'City does not require complex blending, further improving computational efficiency. Fig.~\ref{fig:efficiency_figure} compares the time consumption of the two methods under the same instruction across different city sizes. Time Consumption is defined as the total time to generate a city of a given size, where we enable parallel processing by running two threads simultaneously. As shown in the figure, Yo'City consistently requires less time than SynCity. Notably, Yo'City maintains strong efficiency even without parallelization (43.40 min for a typical $3{\times}3$ city), which is only 69.47\% of SynCity's runtime.
\subsection{Limitations and Future Work}
Yo'City relies on pre-trained models for inference and application. While this approach reduces the dependency on data, allowing for more flexible and free inputs, the overall performance may be influenced by the capabilities of the off-the-shelf models. Therefore, Yo'City should continue to integrate with cutting-edge methods and continually enhance its competence. Additionally, the current model primarily focuses on the overall structure of the city and its infrastructure, without considering natural environmental factors surrounding the urban area, such as mountains, seas, and other geographical features. Future research could look into incorporating these elements to further improve the model’s ability.

\clearpage
\begin{figure*}[t]
\centering
\begin{tcolorbox}[
    colback=gray!5,
    colframe=gray!60,
    title=\textbf{Prompt of Global Planner},
    fonttitle=\bfseries,
    width=\textwidth,   
    boxrule=0.5pt
]
\small
You are helping to design a 3D urban environment based on a textual scene description. The task overview is as follows:

\textbf{Determine Layout Size}: First, decide the overall city layout size in a grid format, such as 2×2, 2×3, or 3×3. The grid represents square sections of the city. The first number is the number of rows, and the second number is the number of columns. Use 2×3 as the default layout. If the scene is large or complex, use 3×3. If the grid layout is not specified in the input, determine it based on the scene description. If a grid layout is already provided, skip this step.

\textbf{Plan and Allocate Areas}: Plan the areas that should be included in the city based on the city instructions, and then allocate these areas to the grid map you determined earlier. 

\textbf{Grid Indexing Rule (Row-major Indexing)}: Each cell has a unique numeric index based on row-first order. For example, in a 2×3 grid:  
(Row 1, Column 1) → 1;
(Row 1, Column 2) → 2; 
(Row 1, Column 3) → 3; 
(Row 2, Column 1) → 4. 

An area can occupy one or multiple cells (for example, a large residential district could span [1, 2, 4, 5]).

\textbf{Output Format}: Output a JSON structure describing the entire city layout and appearance. The JSON should start by specifying the grid size, followed by the list of defined areas. The structure must include:

\begin{itemize}
    \item \textbf{"Grid Size"} — specify the layout as "rows × columns", for example "2×3".
    \item \textbf{"Areas"} — a collection of city areas. Each area should include:
    \begin{itemize}
        \item "Area Name" — the name or type of the area (for example, "Residential Zone", "Commercial Center", "Industrial Zone").
        \item "Description" — a rich and detailed explanation of the area, including:
        \begin{itemize}
            \item building types and architectural styles
            \item atmosphere or functionality (dense, modern, industrial, mixed-use, etc.)
        \end{itemize}
        \item "Grid Index" — a list of grid cells occupied by this area.
    \end{itemize}
\end{itemize}

\textbf{Output Example}:  
\begin{lstlisting}
{
    "Grid Size": "1 X 3",
    "Areas": {
        "Residential District": {
            "Description": "A medium-density housing zone with 4-6 story apartment buildings, internal courtyards, and narrow streets. Buildings are arranged in blocks with small plazas and parking areas.",
            "Grid Index": [1, 2]
        },
        "Commercial Center": {
            "Description": "A bustling commercial core with multi-story malls, office towers, and cafes. The streets are wide and intersect at a central avenue connecting to nearby residential areas.",
            "Grid Index": [3]
        }
    }
}
\end{lstlisting}

\textbf{Important Notes}:
\begin{enumerate}
    \item Focus primarily on generating areas with buildings and city infrastructure. 
    \item When generating parks or plazas, they must:
    \begin{enumerate}
        \item be integrated with built environments (e.g., surrounded by office towers, cafés, residential blocks)
        \item include urban details such as paths, benches, fountains, or sculptures
        \item serve as functional public spaces within the city context
    \end{enumerate}
    \item While ensuring the overall comprehensiveness of urban design and meeting user needs, appropriate additions such as entertainment areas, shopping zones, and cultural and recreational districts can be considered.
    \item Make the descriptions vivid, realistic, and spatially logical — suitable for 3D city modeling. Avoid generic phrases; Describe key visual and structural features. Use coherent relationships between adjacent grid cells (for example, commercial zones near transport hubs, industrial areas near city edges, residential zones adjacent to commercial areas).
\end{enumerate}
\vspace{1mm}

\textcolor{myblue}{City Instruction}:

\textcolor{myblue}{Reference City Summary (Optional):}
\end{tcolorbox}
\captionsetup{labelformat=empty}
\caption{}
\label{Global_Planner}
\end{figure*}
\clearpage
\begin{figure*}[t]
\centering
\begin{tcolorbox}[
    colback=gray!5,
    colframe=gray!60,
    title=\textbf{Prompt of Local Designer},
    fonttitle=\bfseries,
    width=\textwidth,   
    boxrule=0.5pt
]
\small
You are helping to generate detailed scene descriptions for text-to-image generation based on a pre-defined 3D city layout. Task Overview is as follows:

\textbf{You will receive:}
\begin{itemize}
    \item The overall city planning instructions (which define the city type, architectural style focus, and general layout rules).
    \item One specific area from that plan, including: Area Name; Area Description; Grid Index (a list of grids that this area occupies).
\end{itemize}

Your task is to create detailed, vivid descriptions for each grid in this area. Each grid should correspond to one description entry. 

Each description should include:
\begin{itemize}
    \item The dominant building types (residential, commercial, office, industrial, etc.).
    \item The general building scale and form (low-rise, mid-rise, high-rise, tower-like, etc.), but avoid giving specific height or floor numbers.
    \item Architectural styles and materials (glass facade, concrete, brick, steel, etc.).
    \item Spatial and structural layout (street grid, building clusters, plazas, intersections, or inner courtyards).
    \item Density and spatial organization (compact, open, uniform, or mixed-use).
    \item Optional architectural or urban details (bridges, rooftop elements, signage, entrances, etc.).
\end{itemize}

\textbf{Scene Requirements}:
\begin{itemize}
    \item When a city has a specific style requirement, make sure to emphasize that style in the description of each grid.
    \item All scenes should represent daytime environments.
    \item For modern urban scenes, the residual buildings should be mid-rise to high-rise structures.
    \item Do not include light or shadow descriptions.
    \item Do not include people, vehicles, or traffic.
    \item After designing buildings, you can also include parks, fountains, plazas, and other urban facilities to make the scene more lively.
    \item Keep focus entirely on architectural, structural, and urban form elements.
    \item Maintain objectivity and spatial coherence suitable for 3D city scene generation.
    \item If the area covers multiple grids, the grids can share a consistent architectural style but differ slightly in layout or function (for example, one grid may contain offices while another extends the same district with commercial buildings or courtyards).
\end{itemize}

\textbf{Output Format:}  
Output a JSON structure where each key is the grid index, and each value is a detailed scene description of that grid.  
Each description should include:
\begin{itemize}
    \item The dominant building types.
    \item The general building scale and form.
    \item Architectural styles and materials.
    \item Spatial and structural layout.
    \item Density and spatial organization.
    \item Optional architectural or urban details.
\end{itemize}

\textbf{Output Example:}
\begin{lstlisting}
{
  "1": "A cluster of mid-rise residential buildings with light gray concrete facades and subtle beige accents around the balconies. The structures form rectangular blocks aligned along an orderly street grid. The overall tone is neutral but varied, with muted stone and concrete textures creating a balanced, realistic appearance. Building spacing is uniform, with separation between clusters.",
  "2": "A continuation of the residential district featuring taller and denser buildings of similar architectural style. Facades combine pale concrete with soft warm tones, such as beige and off-white, maintaining visual harmony while avoiding monotony. The layout emphasizes a strong linear arrangement along a central avenue, preserving consistency in material and color palette throughout the district."
}
\end{lstlisting}

\textbf{Important Notes:}
\begin{itemize}
    \item Focus on architectural features.
    \item Keep descriptions consistent and technically clear, avoiding unnecessary embellishment.
    \item Ensure each grid description is spatially coherent and realistic, suitable for 3D city generation or text-to-image modeling.
\end{itemize}
\vspace{1mm}
\textcolor{myblue}{City Instruction:} 

\textcolor{myblue}{Area Description and Grid Indices:}
\end{tcolorbox}
\captionsetup{labelformat=empty}
\caption{}
\label{Local_Designer}
\end{figure*}

\clearpage
\begin{figure*}[t]
\centering
\begin{tcolorbox}[
    colback=gray!5,
    colframe=gray!60,
    title=\textbf{Prompt for Generating Image},
    fonttitle=\bfseries,
    width=\textwidth,   
    boxrule=0.5pt
]
\small
You are an expert AI image generator specializing in realistic architectural visualization and urban design. You are generating \{\texttt{{city\_instruction}}\}, which should be the global context.  
Your task is to generate high-resolution, photorealistic, dynamic, properly colorful but harmonious isometric cityscapes based on user-provided base platforms or layouts.  

\textbf{All generations must follow these core principles:}
\begin{enumerate}
    \item The generated scene must strictly remain within the boundaries of the provided square or rectangular platform. Use the base only for spatial confinement — architectural tone and materials should be fully independent.
    \item Emphasize realistic materials, accurate spatial layout, and diverse architectural forms.
    \item Avoid empty or underdeveloped areas — maintain a balanced but not overcrowded density of buildings. 5 to 6 buildings are recommended for a square platform.
    \item The buildings within each area should share a consistent architectural style and overall visual identity — for example, similar materials, color palettes, or design language. However, they should not look identical. Each building should have subtle variations in features such as height, width, facade design, or roof shape, to create a natural and realistic diversity within the same stylistic family.
    \item Ensure all buildings are distinct, non-overlapping, and harmoniously distributed.
    \item The colors can be made a bit richer and more vivid, but avoid excessive saturation.  
    \begin{itemize}
        \item Use a diverse yet harmonious color palette — incorporate complementary and natural tones with moderate contrast between buildings. Each structure should show subtle variations in hue and material (e.g., brick, concrete, glass, stone, or wood), ensuring visual richness without breaking overall unity. Avoid monotone or oversaturated appearances. Glass buildings can be viewed as blue.
        \item Each building has better feature distinct yet coordinated colors — soft warm and cool tones mixed together. Include light beige, terracotta, muted teal, pale yellow, stone gray, and slate blue for a balanced but colorful palette.
    \end{itemize}
    \item Absolutely no letters, logos, words, or recognizable signage on any structures.
    \item Use isometric or slightly elevated perspective to show the entire layout clearly.
    \item No shadows, lighting effects, or atmospheric haze — keep uniform neutral lighting.
    \item Do not generate any people, crowds, or vehicles. Do not generate too much trees and lawns. The root of trees should be thick.
    \item Do not include logos or similar elements in the description.
    \item The output must look realistic, clean, and visually aesthetic.
    \item The entire scene must fit within the visible square platform without external extensions.
\end{enumerate}
\vspace{1mm}
\textcolor{myblue}{City Instruction:}

\textcolor{myblue}{Grid Description:}
\end{tcolorbox}
\vspace{-4mm}
\captionsetup{labelformat=empty}
\caption{}
\label{Image_Generator_1}
\end{figure*}

\begin{figure*}[t]
\centering
\begin{tcolorbox}[
    colback=gray!5,
    colframe=gray!60,
    title=\textbf{Prompt for Refining Image},
    fonttitle=\bfseries,
    width=\textwidth,   
    boxrule=0.5pt
]
\small
You are an expert AI system specialized in architectural visualization and image editing.  
Your goal is to generate or modify images into clean, realistic isometric cityscapes with a pure white background.  

\textbf{Follow these core rules for all outputs:}
\begin{enumerate}
    \item The output must maintain an isometric perspective consistent with the original reference.
    \item For areas which are not residual districts, assess whether the buildings display insufficient architectural diversity. If diversity is low, enhance it in a controlled and realistic way by subtly varying each building’s height, width, roof geometry, façade articulation, and material texture (adding some different logos is also OK). These adjustments should introduce clear visual distinction among buildings while maintaining the original count, layout, spacing, and isometric perspective exactly as in the reference image. Do not alter their relative positions or the overall massing composition. The modifications must remain structurally plausible and stylistically coherent — each building should still clearly correspond to its original form and footprint, yet possess a unique architectural identity through nuanced differences in proportion, façade pattern, tone, and reflectivity. The color tones should be harmonious and consistent.
    \item Completely remove all ground-related elements — including any bases, platforms, tiles, or other floor structures.
    \item The entire background must be pure white (\#FFFFFF) with no visible surface, ground, or shadows.
    \item Only preserve the main and important objects, such as skyscrapers, residential buildings, museums, libraries, theaters, cultural plazas, sculptures, and trees.
    \item Ensure every building has different appearances and colors. But keep the overall style harmonious.
    \item If the image has too many trees or lawns, remove some of them. And make the roots of trees thicker. For scenarios such as parks, keep the trees and lawns.
    \item Remove any incomplete, cut-off, or deformed buildings and objects.
    \item Delete small, cluttered, redundant, or heavily obscured items to keep the composition clear and balanced.
    \item Ensure all objects are distinct and properly spaced — no overlaps, intersections, or unrealistic blending between elements.
    \item Slight adjustments to the appearance, position, or proportion of buildings are allowed to enhance realism and aesthetic balance, but the overall layout and isometric view must remain consistent.
    \item Do not retain any people, crowds, vehicles, or text.
    \item The final image should look clean, realistic, dynamic, and visually harmonious, showcasing an environment on a pure white background.
\end{enumerate}
\vspace{1mm}
\textcolor{myblue}{[Previously Generated Image]}
\end{tcolorbox}
\captionsetup{labelformat=empty}
\caption{}
\label{Image_Generator_2}
\end{figure*}

\begin{figure*}[t]
\centering
\begin{tcolorbox}[
    colback=gray!5,
    colframe=gray!60,
    title=\textbf{Prompt for Evaluating Generated Image},
    fonttitle=\bfseries,
    width=\textwidth,   
    boxrule=0.5pt
]
\small
You are a professional architectural visualization review system.  
Your task is to evaluate an input image based on the corresponding text-to-image prompt provided by the user.  
Perform a single, objective inspection according to the following criteria:

\textbf{Evaluation Criteria:}
\begin{enumerate}
    \item Check whether the ground, platform or any other floor elements have been completely removed (reasonable ground facilities are allowed). If the ground area is entirely white with no visible tiles or other floor elements, this criterion is considered passed.
    \item Determine whether the scene includes a proper number of buildings (it should not look empty or sparse).  
    \item Ensure the layout is not overcrowded — the density should be balanced and harmonious.  
    \item Verify that buildings do not overlap or intersect unnaturally.  
    \item Confirm there are no excessive small, cluttered, or irrelevant objects that reduce visual clarity.  
    \item Check whether the image accurately matches the user’s provided text-to-image description in both content and style.  
    \item Evaluate whether any buildings appear distorted or structurally abnormal.
\end{enumerate}

\textbf{Evaluation Rules:}
\begin{itemize}
    \item Output format (exactly as shown):
\end{itemize}
\begin{verbatim}
Score: [0–10]
Reason: [short explanation]
Rewrite: [revised text-to-image prompt]
\end{verbatim}

\textbf{Scoring rules:}
Don’t be too strict. Being reasonable is more important.
\begin{itemize}
    \item 10 → Perfectly meets all standards.  
    \item 8–9 → Minor imperfections but overall very good.  
    \item 6-7 → Acceptable but needs improvement.  
    \item 4–5 → Noticeable issues; not acceptable.  
    \item 1–3 → Major flaws or clearly poor match.  
    \item 0 → Really bad case (ambiguity appearance, dirty ground...)
\end{itemize}

\textbf{Special Cases:}
\begin{itemize}
    \item If the score is below 6:
    \begin{itemize}
        \item If the issue is incomplete ground removal → reprint the original prompt unchanged in the "Rewrite" field.  
        \item If the issue concerns density, layout, clutter, or mismatched style → rewrite the text-to-image prompt to better align with the standards above.
    \end{itemize}
\end{itemize}
\vspace{1mm}
\textcolor{myblue}{[Refined Image]}

\textcolor{myblue}{Grid Description:}
\end{tcolorbox}
\captionsetup{labelformat=empty}
\caption{}
\label{Image_Evaluator}
\end{figure*}

\begin{figure*}[t]
\centering
\begin{tcolorbox}[
    colback=gray!5,
    colframe=gray!60,
    title=\textbf{Prompt to Generate Expansion Constraints},
    fonttitle=\bfseries,
    width=\textwidth,   
    boxrule=0.5pt
]
\small
You are an expert AI urban designer and 3D scene planner specializing in boundless city generation and expansion.  
Your task is to design and describe new city grid blocks that seamlessly integrate into an existing large-scale city layout.

\textbf{You will receive:}
\begin{itemize}
    \item A rendered image of the current city (top-down or isometric).
    \item A list of existing city zones with their names and brief descriptions.
    \item A user request specifying a new block to add (e.g., "Middle High School", "Tech Innovation Campus") and its grid position.
\end{itemize}

\textbf{Your objectives:}
\begin{enumerate}
    \item Analyze the existing city’s architectural logic, density, and functional organization.
    \item Design a new grid block that visually, structurally, and thematically harmonizes with the current layout.
    \item Provide a concise yet expressive description focused purely on architecture and spatial structure, not atmosphere or storytelling.
\end{enumerate}

\textbf{Architectural Description Rules}
\begin{itemize}
    \item Focus on buildings, massing, form, facades, courtyards, and layout rhythm.
    \item Do not describe people, vehicles, lighting, or atmosphere.
    \item Avoid mentioning time of day, shadows, or weather.
    \item Avoid excessive greenery; mention trees or vegetation only if architecturally essential.
    \item Emphasize proportional balance, hierarchy, material consistency, and spatial continuity.
    \item The district can have multiple buildings (5 to 6 is better).
\end{itemize}

\textbf{Spatial Relation Rules}
When describing spatial relations, consider functional adjacency, visual continuity, and commuting convenience between the new block and existing zones.
\begin{itemize}
    \item Evaluate how easily one can move between them, or how their functions complement each other.
    \item Use conceptual proximity terms only: \texttt{"near"}, \texttt{"relatively\_near"}, \texttt{"slightly\_near"}, \texttt{"far"}.
    \item If two regions do not exhibit a clear spatial relationship, no edge is generated between them.
    \item The \texttt{"far"} relation is only assigned in cases of explicit spatial separation, such as between industrial and residential areas, rather than being applied broadly.
    \item Ensure your spatial relationships are logically consistent with the described city structure (e.g., a school should not be \texttt{"near"} an industrial plant if that contradicts planning logic).
    \item List only meaningful relations; omit irrelevant zones.
\end{itemize}

\textbf{Output Format (JSON only)}
\begin{lstlisting}
{
    "block_name": "<short descriptive name of the new block>",
    "block_description": "<120 to 200 word architectural description focusing on structure and layout>",
    "spatial_relations": {
        "<existing_zone_name>": "<near | relatively_near | slightly_near | far>"
    }
}
\end{lstlisting}

\textbf{Example Output:}
\begin{lstlisting}
{
    "block_name": "Middle High School Block",
    "block_description": "This educational grid introduces a cohesive academic campus composed of multiple wings arranged around a central courtyard. The main teaching hall aligns with the city's grid axis, creating clear pedestrian circulation. Its architecture favors geometric order and rhythmic facades, with restrained material palettes of glass and stone. The overall layout ensures a balanced skyline and coherent integration with nearby urban functions.",
    "spatial_relations": {
        "Urban Residential District": "near",
        "Central Business District": "relatively_near"
    }
}
\end{lstlisting}
\vspace{1mm}
\textcolor{myblue}{[Render Image of Current City]}

\textcolor{myblue}{City Overview:}

\textcolor{myblue}{Expansion Preference:}
\end{tcolorbox}
\captionsetup{labelformat=empty}
\caption{}
\label{Relationship_Constraints}
\end{figure*}

\end{document}